\pdfoutput=1

\documentclass[11pt]{article}

\usepackage{acl}

\usepackage{times}
\usepackage{latexsym}

\usepackage[T1]{fontenc}

\usepackage[utf8]{inputenc}

\usepackage{microtype}

\usepackage{graphicx}
\usepackage{booktabs}
\usepackage{amsthm,amsmath,amssymb,amssymb,bm,mathrsfs}
\usepackage{textcomp}
\usepackage{xcolor}
\usepackage{changes}
\usepackage{algorithm}
\usepackage[noend]{algpseudocode}
\usepackage{multirow}

\usepackage{listings}
\usepackage{changes}

\usepackage{balance}
   
\usepackage[symbol]{footmisc}
\title{Perceiving the World: Question-guided Reinforcement Learning for Text-based Games}

\author{Yunqiu Xu$^1$, Meng Fang$^2$\Thanks{~Corresponding author}, Ling Chen$^1$, Yali Du$^3$, Joey Tianyi Zhou$^4$, Chengqi Zhang$^1$\\
$^1$
University of Technology Sydney, Sydney, Australia \\
$^2$Eindhoven University of Technology, Eindhoven, the Netherlands \\
$^3$King’s College London, London, United Kingdom \\
$^4$  A*STAR Centre for Frontier AI Research (CFAR), Singapore \\
\texttt{\{Yunqiu.Xu,Ling.Chen,Chengqi.Zhang\}@uts.edu.au}, \texttt{m.fang@tue.nl} \\
\texttt{yali.du@kcl.ac.uk}, \texttt{zhouty@ihpc.a-star.edu.sg}\\
}

\begin{document}
\maketitle
\begin{abstract}

Text-based games provide an interactive way to study natural language processing. 
While deep reinforcement learning has shown effectiveness in developing the game playing agent, the low sample efficiency and the large action space remain to be the two major challenges that hinder the DRL from being applied in the real world.
In this paper, we address the challenges by introducing world-perceiving modules, which automatically decompose tasks and prune actions by answering questions about the environment. 
We then propose a two-phase training framework to decouple language learning from reinforcement learning, which  further improves the sample efficiency.
The experimental results show that the proposed method significantly improves the performance and sample efficiency. Besides, it shows robustness against compound error and limited pre-training data.

\end{abstract}

\section{Introduction}

Text-based games are simulated environments where the player observes textual descriptions, and acts using text commands~\cite{hausknecht2019jericho,urbanek2019light}.
These games provide a safe and interactive way to study natural language understanding, commonsense reasoning, and dialogue systems.
Besides language processing techniques, Reinforcement Learning has become a quintessential methodology for solving text-based games. 
Some RL-based game agents have been developed recently and proven to be effective in handling challenges such as language representation learning and partial observability~\cite{narasimhan2015language,fang2017learning,ammanabrolu2019kgdqn}.

Despite the effectiveness, there are two major challenges for RL-based agents, preventing them from being deployed in real world applications: the \textit{low sample efficiency}, and the \textit{large action space}~\cite{dulac2021challenges}. 
The low sample efficiency is a crucial limitation of RL which refers to the fact that it typically requires a huge amount of data to train an agent to achieve human-level performance~\cite{tsividis2017human}.
This is because human beings are usually armed with prior knowledge so that they don't have to learn from scratch~\cite{dubey2018human}.
In a language-informed RL system, in contrast, the agent is required to conduct both language learning and decision making regimes, where the former can be considered as prior knowledge and is much slower than the later~\cite{hill2021grounded}. 
The sample efficiency could be improved through pre-training methods, which decouple the language learning from decision making~\cite{su2017sample}. 
The selection of pre-training methods thus plays an important role: if the pre-trained modules perform poorly on unseen data during RL training, the incurred compound error will severely affect the decision making process. 
Another challenge is the large discrete action space: the agent may waste both time and training data if attempting irrelevant or inferior actions~\cite{dulac2015actionspace,zahavy2018nips}. 

In this paper, we aim to address these two challenges for reinforcement learning in solving text-based games. 
Since it is inefficient to train an agent to solve complicated tasks (games) from scratch, we consider decomposing a task into a sequence of subtasks as inspired by~\cite{andreas2017modular}.
We design an RL agent that is capable of automatic task decomposition and subtask-conditioned action pruning, which brings two branches of benefits.
First, the subtasks are easier to solve, as the involved temporal dependencies are usually short-term.
Second, by acquiring the skills to solve subtasks, the agent will be able to learn to solve a new task more quickly by reusing the learnt skills~\cite{barreto2020fast}. 
The challenge of large action space can also be alleviated, if we can filter out the actions that are irrelevant to the current subtask. 

\begin{figure*}[t!]
\centering
\includegraphics[width=1.0\textwidth]{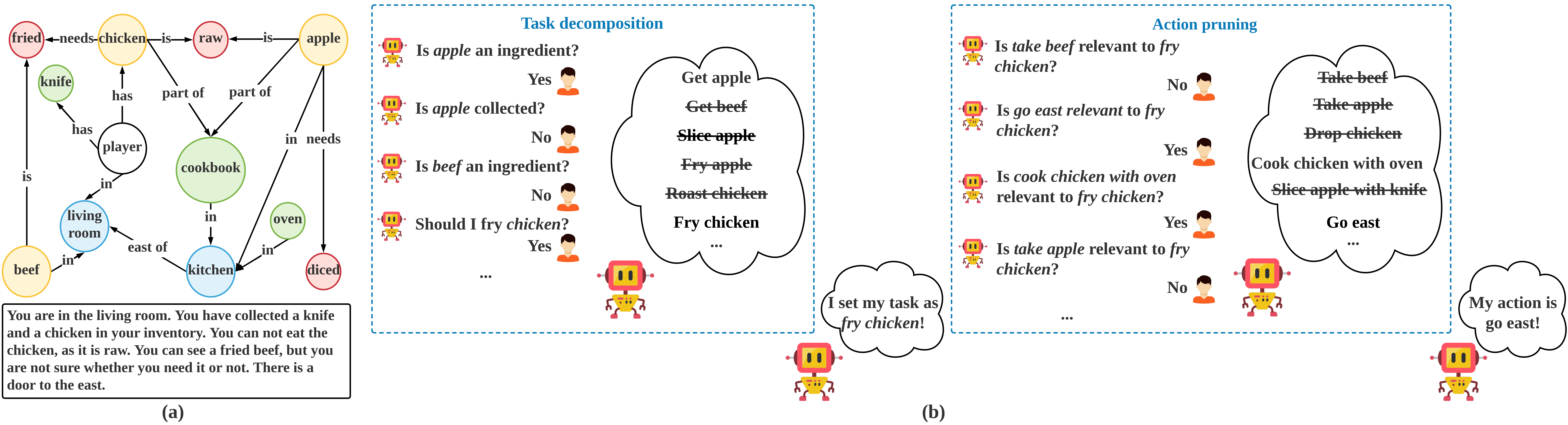}
\caption{(a) An example of the observation, which can be textual, KG-based, or hybrid. (b) The decision making process. Through question answering, the agent is guided to first decompose the task as subtasks, then reduce the action space conditioned on the subtask.}
\label{arch_1_overview}
\end{figure*}

Inspired by the observation that human beings can understand the environment conditions through question answering~\cite{das2020probing,ammanabrolu2020kga2cexplore}, we design world-perceiving modules to realize the aforementioned functionalities (i.e., task decomposition and action pruning) and name our method as \textbf{Q}uestion-guided \textbf{W}orld-perceiving \textbf{A}gent (QWA)\footnote{Code is available at: \url{https://github.com/YunqiuXu/QWA}}.
Fig. \ref{arch_1_overview} (b) shows an example of our decision making process. 
Being guided by some questions, the agent first decomposes the task to obtain a set of available subtasks, and selects one from them. Next, conditioned on the selected subtask, the agent conducts action pruning to obtain a refined set of actions. 
In order to decouple language learning from decision making, which further improves the sample efficiency, we propose to acquire the world-perceiving modules through supervised pre-training. 
We design a two-phase framework to train our agent. 
In the first phase, a dataset is built for the training of the world-perceiving modules. 
In the second phase, we deploy the agent in games with the pre-trained modules frozen, and train the agent through reinforcement learning.

We conduct experiments on a series of cooking games. We divide the games as simple games and complex games, and construct the pre-training dataset from simple games only.
The experimental results show that QWA achieves high sample efficiency in solving complex games. We also show that our method enjoys robustness against compound error and limited pre-training data. 

Our contributions are summarized as follows:
Firstly, we develop an RL agent featured with question-guided task decomposition and action space reduction. 
Secondly, we design a two-phase framework to efficiently train the agent with limited data.
Thirdly, we empirically validate our method's effectiveness and robustness in complex games. 

\section{Related work}

\subsection{RL agents for text-based games}

The RL agents for text-based games can be divided as text-based agents and KG-based agents based on the form of observations. 
Compared with the text-based agents~\cite{narasimhan2015language,yuan2018counting,adolphs2019ledeepchef,jain2019aaai,yin2019cog,xu2020cog,guo2020emnlp}, which take the raw textual observations as input to build state representations, the KG-based agents construct the knowledge graph and leverage it as the additional input~\cite{ammanabrolu2019kgdqn,xu2020nips}.
By providing structural and historical information, the knowledge graph helps the agent to handle partial observability, reduce action space, and improve generalizability across games.
Based on how actions are selected, the RL agents can also be divided as parser-based agents, choice-based agents, and template-based agents. 
The parser-based agents generate actions word by word, leading to a huge combinatorial action space~\cite{kohita2021language}. 
The choice-based agents circumvent this challenge by assuming the access to a set of admissible actions at each game state~\cite{he2016drrn}.
The template-based agents achieve a trade-off between the huge action space and the assumption of admissible action set by introducing the template-based action space, where the agent selects first a template, and then a verb-object pair either individually~\cite{hausknecht2019jericho} or conditioned on the selected  template~\cite{ammanabrolu2019kga2c}.
In this work, we aim to improve the sample efficiency and reduce the action space through pre-training. Being agnostic about the form of observations and the action selecting methods, our work complements the existing RL agents. 

\subsection{Hierarchical RL}

Our work is closely related to task decomposition~\cite{oh2017zero,shiarlis2018taco,sohn2018subtask} and hierarchical reinforcement learning~\cite{dayan1992hrl_feudal,kulkarni2016hrl,vezhnevets2017hrl_feudal}.
Similar to our efforts, ~\citet{jiang2019hrl} and \citet{xu2021emnlp} designed a meta-policy for task decomposition and subtask selection, and a sub-policy for goal-conditioned decision making. 
Typically, these works either assume the access to a set of available subtasks, or decompose a task through pre-defined rules, while we aim to achieve automatic task decomposition through pre-training, and remove the requirement for expert knowledge during reinforcement learning. 
Besides, existing work assumes that unlimited interaction data can be obtained to train the whole model through RL. In contrast, we consider the more practical situation where the interaction data is limited, and focus on improving the RL agent's data efficiency. 
Regarding the sub-policy, we do not assume the access to the termination states of the subtasks. We also do not require additional handcrafted operations in reward shaping~\cite{bahdanau2018goal_language}.

\subsection{Pre-training methods for RL}

There have been a wide range of work studying pre-training methods or incorporating pre-trained modules to facilitate reinforcement learning~\cite{eysenbach2018pretraining,hansen2019pretraining,sharma2019pretraining,gehring2021pretraining,liu2021pretraining,schwarzer2021pretraining}. 
One major branch among them is Imitation Learning (IL), where the agent is trained to imitate human demonstrations before being deployed in RL~\cite{hester2018pretraining,zhu2018pretraining,reddy2019pretraining}. 
Although we also collect human labeled data for pre-training, we leverage the data to help the agent to perceive the environment instead of learning the solving strategies. Therefore, we do not require the demonstrations to be perfect to solve the game. 
Besides, our method prevails when pre-trained on simple tasks rather than complicated ones, making it more feasible for human to interact and annotate~\cite{arumugam2017annotating,mirchandani2021ella}. Further discussions to compare our method with IL are provided in subsequent sections. 

In the domain of text-based games, some prior works have involved pre-training tasks such as state representation learning~\cite{ammanabrolu2020quest,singh2021pretraining}, knowledge graph constructing~\cite{murugesan2020commonsense} and action pruning~\cite{hausknecht2019nail,tao2018towards,yao2020emnlp}. 
For example, ~\citet{ammanabrolu2020kga2cexplore} designed a module to extract triplets from the textual observation by answering questions, and use these triplets to update the knowledge graph. 
As far as we know, we are the first to incorporate pre-training based task decompositon in this domain. 
Besides, instead of directly pruning the actions based on the observation, we introduce subtask-conditioned action pruning to further reduce the action space.

\section{Background}
\paragraph{POMDP} Text-based games can be formulated as a Partially Observable Markov Decision Processes (POMDPs) \cite{cote2018textworld}. A POMDP can be described by a tuple $\mathcal{G} =\langle \mathcal{S}, \mathcal{A}, P, r, \Omega, O, \gamma \rangle$, with $\mathcal{S} $ representing the state set, $\mathcal{A}$ the action set, $P(s'|s,a):\mathcal{S}\times \mathcal{A}\times \mathcal{S}\mapsto\mathbb{R}^+$ the state transition probabilities, $r(s,a):\mathcal{S}\times \mathcal{A}\mapsto\mathbb{R}$ the reward function,  $\Omega$ the observation set,  $O$ the conditional observation  probabilities, and $\gamma \in (0,1]$ the discount factor. 
At each time step, the agent receives an observation $o_t \in \Omega$ based on the probability $O(o_t|s_t,a_{t-1})$, and select an action $a_t \in \mathcal{A}$.
The environment will transit into a new state based on the probability $T(s_{t+1}|s_t,a_t)$, and return a scalar reward $r_{t+1}$. 
The goal of the agent is to select the action to maximize the expected cumulative discounted rewards:  $R_t = \mathbb{E}[\sum_{t=0}^{\infty}\gamma^k r_{t}]$.  

\paragraph{Observation form} In text-based games, the observation can be in the form of text, knowledge graph, or hybrid.
Fig. \ref{arch_1_overview} (a) shows an example of the textual observation and the corresponding KG-based observation. 
We do not make assumptions about the observation form and our method is compatible with any of those forms.

\paragraph{Problem setting} 

\begin{figure}[t!]
\centering
\includegraphics[width=0.45\textwidth]{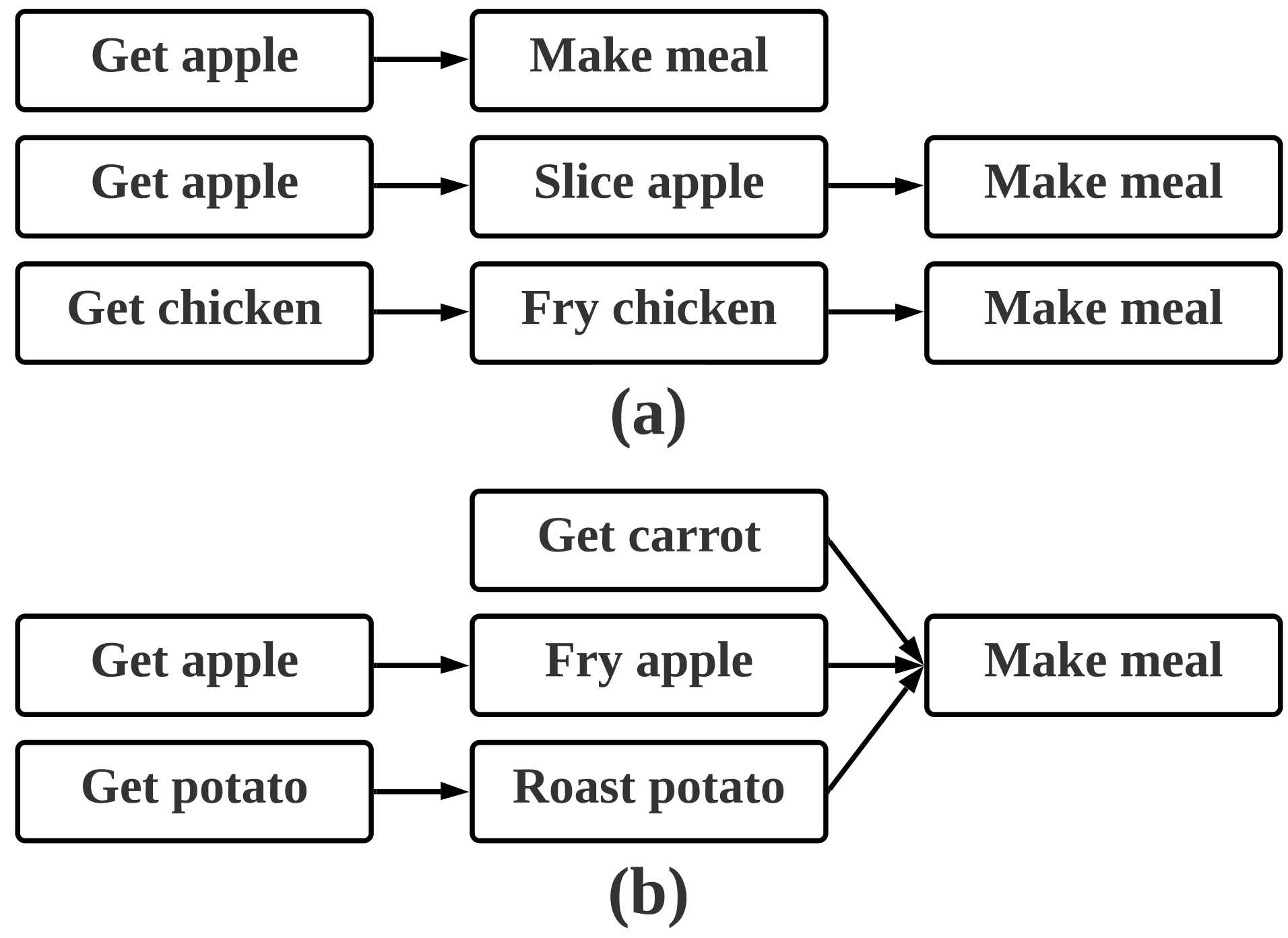}
\caption{Subtasks for solving (a) 3 simple games and (b) 1 complex game.}
\label{arch_2_subtask}
\end{figure}

We aim to design an RL-based agent that is able to conduct automatic task decomposition and action pruning in solving text-based games. 
We consider games sharing similar themes and tasks, but varying in their complexities~\cite{adhikari2020gatav2,chen2021askyourhumans}.
Taking the cooking games~\cite{cote2018textworld} as an example, the task is always ``make the meal''. To accomplish this task, the agent has to explore different rooms to collect all ingredients, prepare them in right ways, and make the meal. 
A game's complexity depends on the number of rooms, ingredients, and the required preparation steps.
We define a subtask as a  milestone towards completing the task (e.g., ``get apple'' if ``apple'' is included in the recipe), and a subtask requires a sequence of actions to accomplish (e.g., the agent has to explore the house to find the apple).
A game is considered simple, if it consists of only a few subtasks, and complex if it consists of more subtasks. 
Fig. \ref{arch_2_subtask} gives examples of simple games and complex games. 
While being closer to real world applications, complex games are hard to solve by RL agents because: 
1) it's expensive to collect sufficient human labeled data for pre-training;
2) it's unrealistic to train an RL agent from scratch. 
We therefore focus on agent's sample efficiency and performance on complex games. Our objective is to leverage the labeled data collected from simple games to speed up RL training in complex games, thus obtaining an agent capable of complex games. 
For more details and statistics of the simple / complex games used in our work, please refer to Sec. \ref{exp_experiment_settings}.

\section{Methodology}

\subsection{Framework overview}

\begin{figure*}[t!]
\centering
\includegraphics[width=0.9\textwidth]{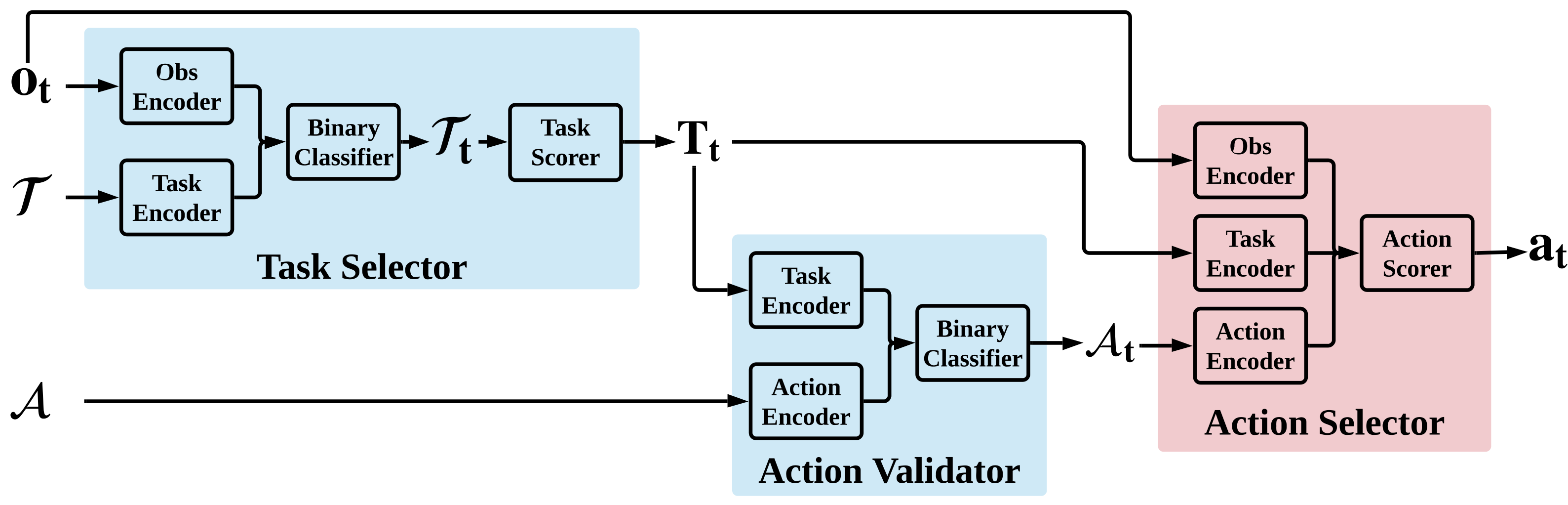}
\caption{The overview of QWA. The blue modules will be trained in the pre-training phase, while the red module will be trained in the RL phase.}
\label{arch_3_agent}
\end{figure*}

Fig. \ref{arch_3_agent} shows the overview of our QWA agent. 
We consider two world-perceiving modules: a task selector and an action validator.
Given the observation $o_t$ and the task candidate set $\mathcal{T}$, we use the task selector to first obtain a subset of currently available subtasks $\mathcal{T}_t \subseteq \mathcal{T}$, then select a subtask $T_t \in \mathcal{T}_t$.
Given $T_t$ and the action candidate set $\mathcal{A}$, we use the action validator to get an action subset $\mathcal{A}_t \subseteq \mathcal{A}$, which contains only those relevant to the subtask $T_t$. 
Finally, given $o_t$ and $T_t$, we use an action selector to score each action $a \in \mathcal{A}_t$, and the action with the highest score will be selected as $a_t$. 

The training of the world-perceiving modules can be regarded as the language learning regime, while the training of the action selector can be regarded as the decision making regime.
We consider a two-phase training strategy to decouple these two regimes to further improve the sample efficiency~\cite{hill2021grounded}.
In the pre-training phase, we collect human interaction data from the simple games, and design QA datasets to train the world-perceiving modules through supervised learning. 
In the reinforcement learning phase, we freeze the pre-trained modules, and train the action selector in the complex games through reinforcement learning.

\subsection{Task selector}

Depending on the experiment settings, $\mathcal{T}$ and $\mathcal{A}$ can be either fixed vocabulary sets (parser-based), or changing over time (choice-based).
We regard a subtask available if it is essential for solving the ``global'' task, and there's no prerequisite subtask. 
For example, the subtask ``get apple'' in Fig. \ref{arch_1_overview}, as the object ``apple'' is an ingredient which has not been collected.
Although another subtask ``dice apple'' is also essential for making the meal, it is not available since there exists a prerequisite subtask (i.e., you should collect the apple before dicing it). 
The aim of the task selector is to identify a subset of available subtasks $\mathcal{T}_t \subseteq \mathcal{T}$, and then select one subtask $T_t \in \mathcal{T}_t$. 

We formulate the mapping $f(o_t, \mathcal{T}) \rightarrow \mathcal{T}_t$ as a multi-label learning problem~\cite{zhang2013multilabellearning}. For simplicity, we assume that the subtask candidates are independent with each other. Thus, the multi-label learning problem can be decomposed as $|\mathcal{T}|$ binary classification problems. 
Inspired by the recent progress of question-conditional probing~\cite{das2020probing}, language grounding~\cite{hill2021grounded}, and QA-based graph construction~\cite{ammanabrolu2020kga2cexplore}, 
we cast these binary classification problems as yes-or-no questions, making the task selector a world-perceiving module. 
For example, the corresponding question for the subtask candidate ``get apple'' could be ``Whether `get apple' is an available subtask?''.
This module can guide the agent to understand the environment conditions through answering questions, but will not directly lead the agent to a specific decision.
We can obtain this module through supervised pre-training, and decouple it from reinforcement learning to yield better sample efficiency. 
Fig. \ref{arch_1_overview} (b) shows some sample QAs, where a human answerer can be replaced by a pre-trained task selector. 

Some previous work also considered task decomposition~\cite{chen2021askyourhumans,hu2019hierarchical}, but the related module is obtained through imitating human demonstrations, which is directly related to decision making instead of world perceiving.
Compared with these work, our method has two folds of benefits.
First, there may exist multiple available subtasks at a timestep. Imitating human demonstrations will specify only one of them, which may be insufficient and lead to information loss. 
Second, we do not require expert demonstrations which guarantee to solve the game. Instead, we can ask humans to annotate either imperfect demonstrations, or even demonstrations from a random agent. We will treat the IL-based method as a baseline and conduct comparisons in the experiments. 

Given the set of available subtasks $\mathcal{T}_t$,  arbitrary strategies can be used to select a subtask $T_t$ from it. For example, we can employ a non-learnable task scorer to obtain $T_t$ by  random sampling, since each subtask $T \in \mathcal{T}_t$ is essential for accomplishing the task. We can also train a task scorer via a meta-policy for adaptive task selection~\cite{xu2021emnlp}.

\subsection{Action validator}

After obtaining the subtask $T_t$, we conduct action pruning conditioned on it (or on both $T_t$ and $o_t$) to reduce the action space,  tackling the challenge of large action space.
Similar to the task selector, we formulate action pruning as $|\mathcal{A}|$ binary classification problems, and devise another world-perceiving module: the action validator.
The action validator is designed to check the relevance of each action candidate $a \in \mathcal{A}$ with respect to $T_t$ by answering questions like ``Is the action candidate `take beef' relevant to the subtask `fry chicken'?'', so as to obtain a subset of actions $\mathcal{A}_t \subseteq \mathcal{A}$ with irrelevant actions filtered. 
Fig. \ref{arch_3_agent} shows the module architecture. 
Similar to the task selector, we pre-train this module through question answering. Sample QAs have been shown in Fig. \ref{arch_1_overview} (b).

\subsection{Action selector \label{method_action_selector}}

After pre-training, we deploy the agent in the complex games, and train the action selector through RL. 
We freeze the pre-trained modules, as no human labeled data will be obtained in this phase. 
At each time step, we use the task selector and the action validator to produce $\mathcal{T}_t$ and $\mathcal{A}_t$, respectively. 
We keep using the same subtask $T$ over time until it is not included in $\mathcal{T}_t$, as we do not want the agent to switch subtasks too frequently. 
The agent can simply treat $T_t$ as the additional observation of $o_t$.
If we do not limit the use of human knowledge in this phase, we can also treat $T_t$ as a goal with either hand-crafted~\cite{jiang2019hrl} or learnt reward function~\cite{colas2020instruction}.
Arbitrary methods can be used for optimizing~\cite{ammanabrolu2019kga2c,adhikari2020gatav2}.

\begin{table*}[t!]
\caption{\label{table_game_statistics} Game statistics. We use the simple games to provide human labeled data in the pre-training phase. We use the medium \& hard games in the reinforcement learning phase.}
\centering
\resizebox{1.0\textwidth}{!}{
\begin{tabular}{c|ccccccccc}
\hline 
\textbf{Name}  & \textbf{Traj.Length} & \textbf{\#Triplets} & \textbf{\#Rooms} & \textbf{\#Objs} & \textbf{\#Ings} & \textbf{\#Reqs} & \textbf{\#Acts} & \textbf{\#Subtasks} & \textbf{\#Avail.Subtasks} \\
\hline
\textbf{Simple} & 7.90 & 38.48 & 5.76 & 23.69 & 1.49 & 0.96 & 14.50 & 12.44 & 1.14\\ 
\textbf{Medium} & 15.30& 51.07 & 6.00 & 26.10 & 3.00 & 3.00 & 23.48 & 23.00 & 1.94\\ 
\textbf{Hard}   & 21.75& 59.95 & 8.00 & 31.48 & 3.00 & 4.00 & 22.94 & 23.00 & 2.16\\ 
\hline
\end{tabular}
}
\end{table*}

One issue we are concerned about is the compound error $-$ the prediction error from imperfect pre-trained modules will adversely affect RL training~\cite{talvitie2014model,racaniere2017i2as}. 
For example, the false predictions made by the binary classifier in the task selector may lead to a wrong $T_t$, which affects $\mathcal{A}_t$ and $a_t$ in turn. 
To alleviate the influence of the compound error, we assign time-awareness to subtasks. A subtask is bounded by a time limit $[0,\xi]$. If the current subtask $T$ is not finished within its time limit, we force the agent to re-select a new subtask $T_t \in \mathcal{T}_t \setminus \{T\}$, regardless whether $T$ is still available. 
Besides making the agent robust against errors, another benefit by introducing time-awareness to subtasks is that it improves the subtask selection diversity, which helps the agent to avoid getting stuck in local minima~\cite{pong2020skewfit,campero2020amigo}.  

\section{Experiments}



\subsection{Experiment settings\label{exp_experiment_settings}}

We conduct experiments on cooking games provided by the rl.0.2 game set\footnote{\url{https://aka.ms/twkg/rl.0.2.zip}} and the FTWP game set\footnote{\url{https://aka.ms/ftwp/dataset.zip}}, which share the vocabulary set. Based on the number of subtasks, which is highly correlated to the number of ingredients \& preparing requirements, we design three game sets with varying complexities: 3488 simple games, 280 medium games and 420 hard games. 
Note that there is no overlapping games between the simple set and the medium / hard game sets. Table \ref{table_game_statistics} shows the game statistics. Besides ``Traj.Length'', which denotes the average length of the expert demonstrations per game\footnote{The demonstrations of the medium \& hard games are just for statistics, and will not be used for pre-training.}, other statistic metrics are averaged per time step per game (e.g., ``\#Subtasks'' and ``\#Avail.Subtasks'' denote the average number of subtask candidates $\mathcal{T}$, and the average number of available subtasks $\mathcal{T}_t$, respectively). 
We will collect human interaction data from the simple games for pre-training. We regard both medium \& hard games as complex, and will conduct reinforcement learning on these two game sets without labeled data.

\subsection{Baselines\label{exp_baselines}}

We consider the following four models, and compare with more variants in ablation studies:
\begin{itemize}
    \item \textbf{GATA}~\cite{adhikari2020gatav2}: a powerful KG-based RL agent, which is the benchmark model for cooking games. 
    \item \textbf{IL}~\cite{chen2021askyourhumans}: a hierarchical agent which also uses two training phases. In the first phase, both the task selector and the action selector are pre-trained through imitation learning. Then in the second phase, the action selector is fine-tuned through reinforcement learning. 
    \item \textbf{IL w/o FT}: a variant of the IL baseline, where only the imitation pre-training phase is conducted, and there's no RL fine-tuning.
    \item \textbf{QWA}: the proposed model with world-perceiving modules.
\end{itemize}

\subsection{Implementation details\label{section_implementation_details}}

\paragraph{Model architecture} All models are implemented based on GATA's released code\footnote{\url{https://github.com/xingdi-eric-yuan/GATA-public}}. In particular, we use the version GATA-GTF, which takes only the KG-based observation, and denote it as GATA for simplicity. The observation encoder is implemented based on the Relational Graph Convolutional Networks (R-GCNs)~\cite{schlichtkrull2018rgcn} by taking into account both nodes and edges. Both the task encoder and the action encoder are implemented based on a single transformer block with single head ~\cite{vaswani2017transformer} to encode short texts. The binary classifier, the task scorer and the action scorer are linear layers. The GATA and IL models are equipped with similar modules. Please refer to Appendix \ref{section_appendix_baseline_details} for details.

\paragraph{Pre-training} We train the task selector and the action validator separately, as they use different types of QAs. 
We ask human players to play the simple games, and answer the yes-or-no questions based on the observations. The details of the dataset construction (interaction data collection, question generation, answer annotation, etc. ) could be found at Appendix \ref{section_appendix_dataset}. 
We train the task selector with a batch size of 256, and the action validator with a batch size of 64. The modules are trained for 10-20 epochs using Focal loss and Adam optimizer with a learning rate of 0.001. 

\paragraph{Reinforcement learning} We consider the medium game set and hard game set as different experiments. We split the medium game set into 200 training games / 40 validation games / 40 testing games, and the hard game set into 300 / 60 / 60. We follow the default setting of ~\cite{adhikari2020gatav2} to conduct reinforcement learning. We set the step limit of an episode as 50 for training and 100 for validation / testing. We set the subtask time limit $\xi=5$. For each episode, we sample a game from the training set to interact with. We train the models for 100,000 episodes. The models are optimized via Double DQN (epsilon decays from 1.0 to 0.1 in 20,000 episodes, Adam optimizer with a learning rate of 0.001) with Pritorized Experience Replay (replay buffer size 500,000). For every 1,000 training episodes, we validate the model and report the testing performance.

\subsection{Evaluation metrics}

We measure the models through their RL testing performance. We denote a game's score as the episodic sum of rewards without discount. 
As different games may have different maximum available scores, we report the normalized score, which is defined as the collected score normalized by the maximum score for a game.

\section{Results and discussions}

\subsection{Main results \label{result_main_results}}

\begin{table}[t!]
\caption{\label{table_main} The testing performance at 20\% / 100\% of the reinforcement learning phase.}
\centering
\resizebox{0.5\textwidth}{!}{
\begin{tabular}{c|cc|cc}
\hline 
\multirow{2}*{\textbf{Model}} & \multicolumn{2}{c}{\textbf{Medium}} & \multicolumn{2}{c}{\textbf{Hard}} \\
 ~ & \textbf{20\%} & \textbf{100\%} & \textbf{20\%} & \textbf{100\%} \\
\hline
\textbf{QWA (ours)} & \textbf{0.66}$\pm$0.02 & \textbf{0.71}$\pm$0.04 & \textbf{0.53}$\pm$0.04 & \textbf{0.53}$\pm$0.02 \\
\textbf{GATA}       & 0.31$\pm$0.02 & 0.57$\pm$0.18 & 0.25$\pm$0.02 & 0.48$\pm$0.01 \\
\textbf{IL}         & 0.45$\pm$0.18 & 0.26$\pm$0.03 & 0.32$\pm$0.11 & 0.35$\pm$0.08 \\
\textbf{IL w/o FT}  
& 0.63$\pm$0.05 & 0.63$\pm$0.05 & 0.48$\pm$0.05 & 0.48$\pm$0.05 \\
\hline
\end{tabular}
}
\end{table}

\begin{figure}[t!]
\centering
\includegraphics[width=0.49\textwidth]{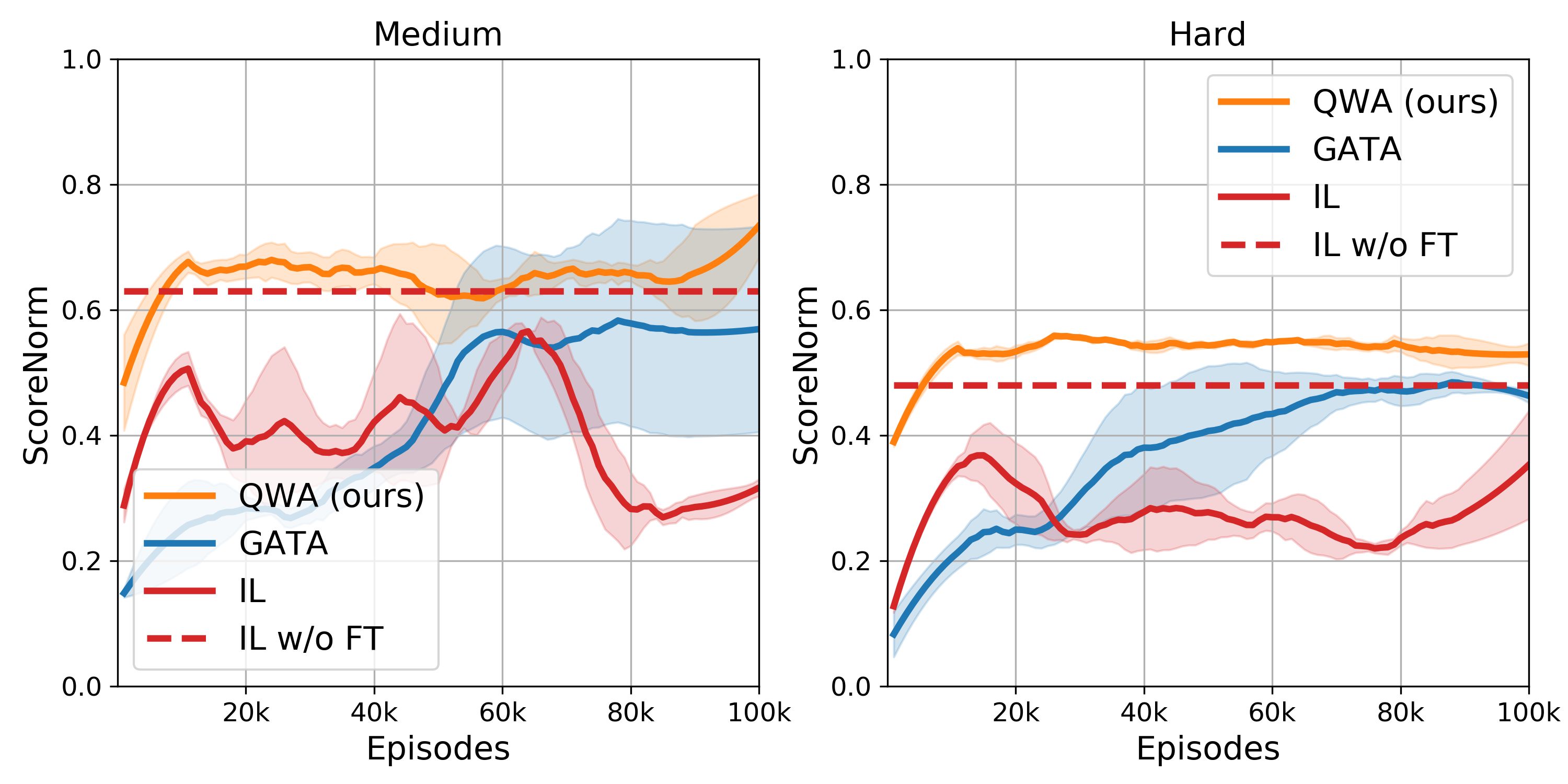}
\caption{The RL testing performance w.r.t. training episodes. The red dashed line denotes the IL agent without fine-tuning.}
\label{exp_1_main}
\end{figure}

Fig. \ref{exp_1_main} shows the RL testing performance with respect to the training episodes. Table \ref{table_main} shows the testing performance after 20,000 training episodes (20\%) / at the end of RL training (100\%).
Compared with GATA, which needs to be ``trained from scratch'', the proposed QWA model achieves high sample efficiency: it reaches convergence with high performance before 20\% of the training stage, saving 80\% of the online interaction data in complex games.
The effectiveness of pre-training can also be observed from the variant ``IL w/o FT'': even though it requires no further training on the medium / hard games, it achieves comparable performance to our model. However, the performance of QWA can be further improved through RL, while it does not work for the IL-based model, as we can observe the performance of ``IL'' becomes unstable and drops significantly during the RL fine-tuning.
A possible reason is that there exists large domain gap between simple and medium (hard) games, and our model is more robust against such domain shifts.
For example, our world-perceiving task selector performs better than IL-based task selector in handling more complex observations (according to Table \ref{table_game_statistics}, the observations in medium / hard games contain more triplets, rooms and objects), facilitating the training of the action selector. 
Besides the domain gap in terms of the observation space, there is also a gap between domains in terms of the number of available subtasks $-$ while there's always one available subtask per time step in simple games, the model will face more available subtasks in the medium / hard games. 
Different from our task selector, which is trained to check the availability of every subtask candidate, the IL pre-trained task selector can not adapt well in this situation, as it is trained to find the unique subtask and ignore the other subtask candidates despite whether they are also available.

\subsection{Performance on the simple games}

We further investigate the generalization performance of our model on simple games, considering that simple games are not engaged in our RL training. To conduct the experiment, after RL training, we deploy all models on a set of 140 held-out simple games for RL interaction. 
Table \ref{table_simple} shows the results, where ``Medium 100\%'' (``Hard 100\%'') denotes that the model is trained on medium (hard) games for the whole RL phase. 
The generalizability of GATA, which is trained purely with medium and hard games, is significantly low and cannot perform well on simple games. 
In contrast, our model performs very well and achieves over 80\% of the scores. The world-perceiving modules, which are pre-trained with simple games, help to train a decision module that adapts well on unseen games. 
It is not surprising that the variant ``IL w/o FT'' also performs well on simple games, since they are only pre-trained with simple games. 
However, as indicated by the performance of ``IL'', after fine-tuning on medium/hard games (recalling Sec. \ref{result_main_results}), the action scorer ``forgets'' the experience/skills dealing with simple games and the model fails to generalize on unseen simple games.
In summary, the best performance achieved by QWA demonstrates that our model can generalize well on games with different complexities.


\begin{table}[t!]
\caption{\label{table_simple} The RL testing performance on simple games.}
\centering
\resizebox{0.38\textwidth}{!}{
\begin{tabular}{c|cc}
\hline 
\textbf{Model} & \textbf{Medium 100\%} & \textbf{Hard 100\%} \\
\hline
\textbf{QWA (ours)} & \textbf{0.80}$\pm$0.01 & \textbf{0.82}$\pm$0.02 \\
\textbf{GATA}       & 0.32$\pm$0.03 & 0.45$\pm$0.12 \\
\textbf{IL}         & 0.44$\pm$0.02 & 0.29$\pm$0.03 \\
\textbf{IL w/o FT}  & 0.76$\pm$0.06 & 0.76$\pm$0.06 \\
\hline
\end{tabular}
}
\end{table}

\subsection{Ablation study}

\begin{figure}[t!]
\centering
\includegraphics[width=0.49\textwidth]{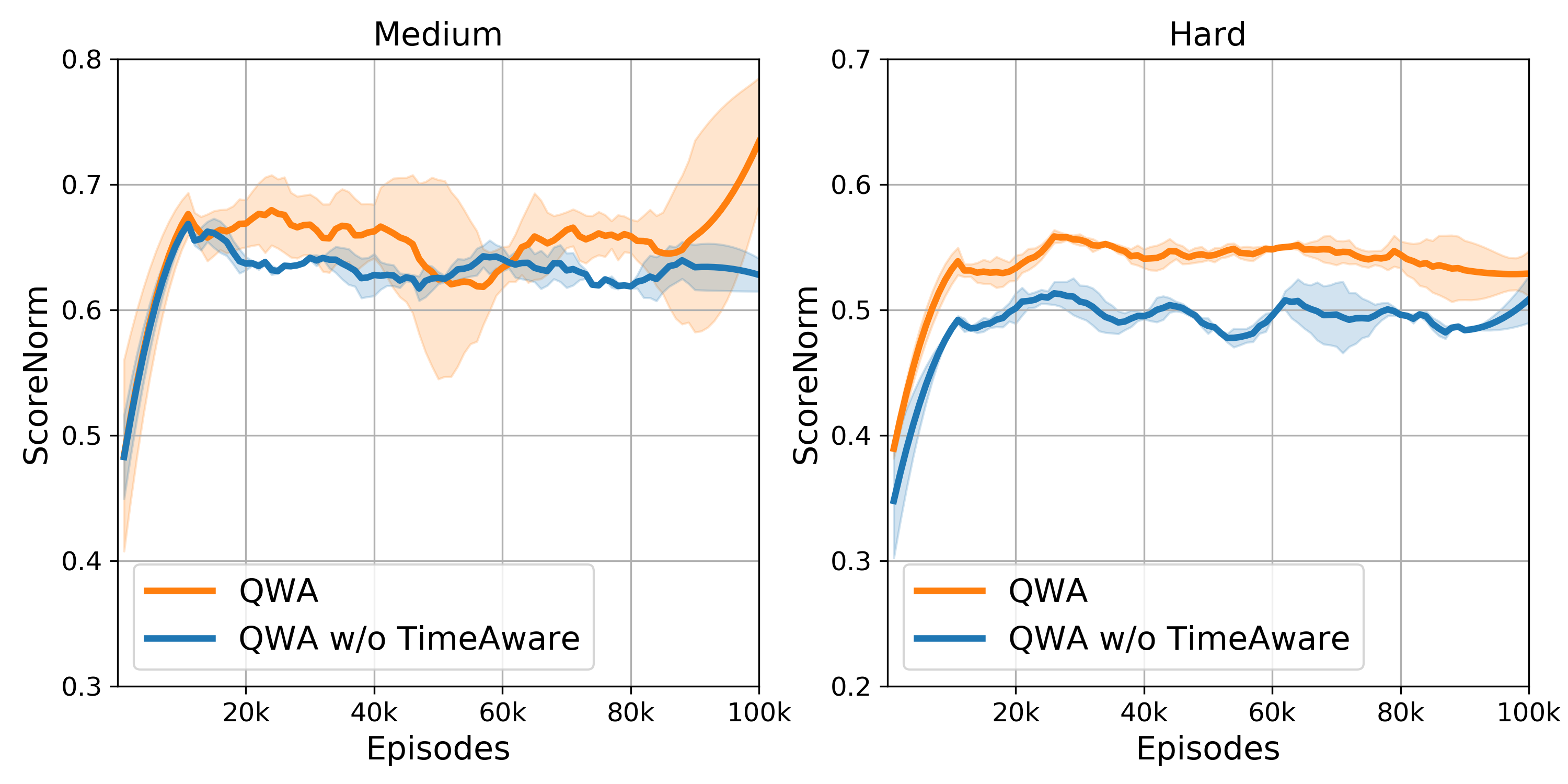}
\caption{The performance of our model and the variant without time-awareness.}
\label{exp_2_ablationTimeAware}
\end{figure}

We study the contribution of the subtask time-awareness by comparing our full model with the variant without this technique. Fig. \ref{exp_2_ablationTimeAware} shows the result. Although the models perform similarly in the medium games, the full model shows better performance in the hard games, where there may exist more difficult subtasks (we regard a subtask more difficult if it requires more actions to be completed). 
Assigning each subtask a time limit prevents the agent from pursuing a too difficult subtask, and improves subtask diversity by encouraging the agent to try different subtasks.
Besides, it prevents the agent from being stuck in a wrong subtask, making the agent more robust to the compound error.

\begin{figure}[t!]
\centering
\includegraphics[width=0.49\textwidth]{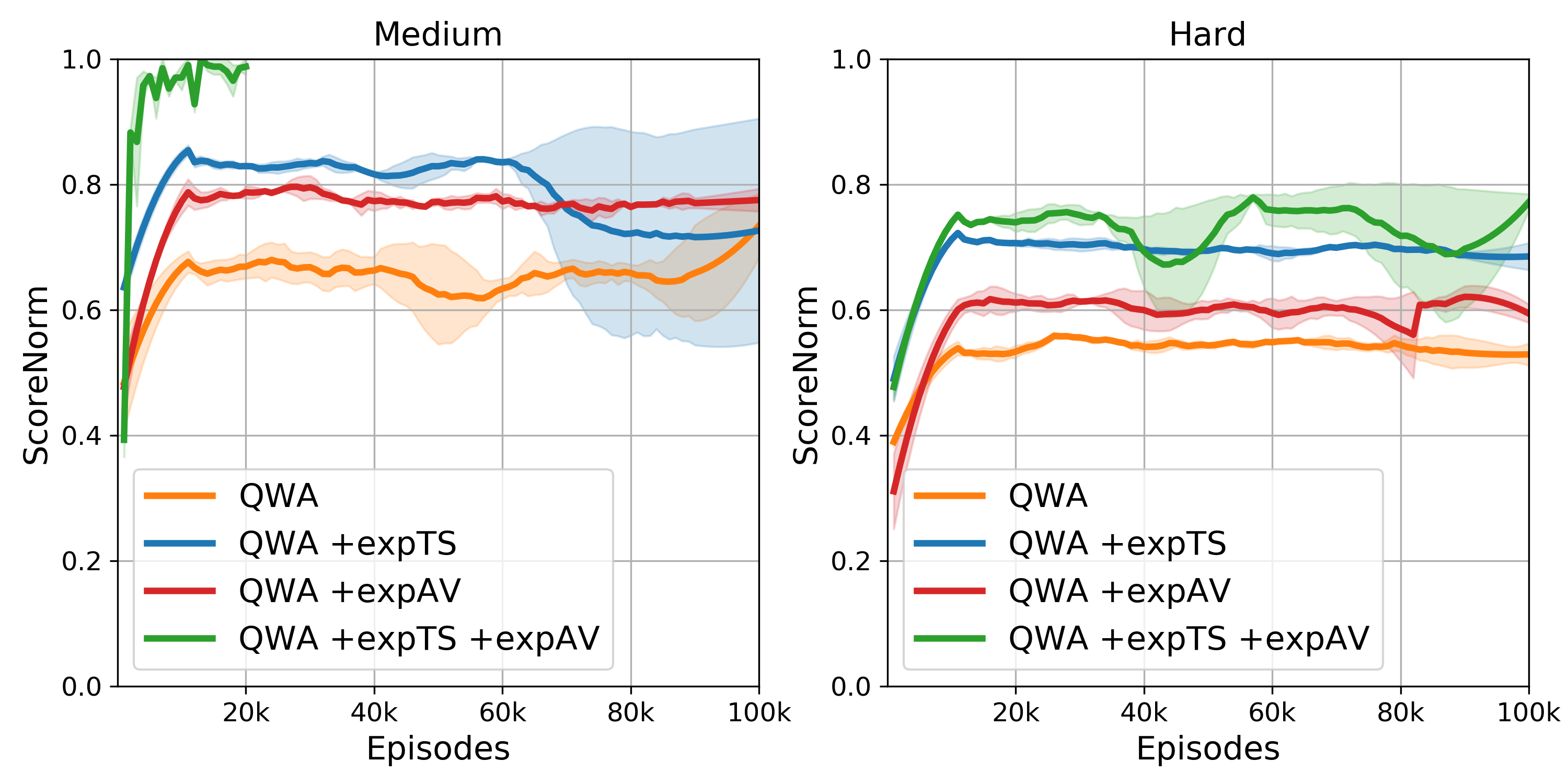}
\caption{The performance of our model and the variants with expert modules.}
\label{exp_2_ablationExpert}
\end{figure}

We then investigate the performance upper bound of our method by comparing our model to variants with oracle world-perceiving modules. 
Fig. \ref{exp_2_ablationExpert} shows the results, where ``+expTS'' (``+expAV'') denotes that the model uses an expert task selector (action validator). 
There's still space to improve the pre-trained modules.
The variant ``QWA +expTS +expAV'' solves all the medium games and achieves nearly 80\% of the scores in hard games, showing the potential of introducing world-perceiving modules in facilitating RL. 
We also find that assigning either the expert task selector or the expert action validator helps to improve the performance.
In light of these findings, we will consider more powerful pre-training methods as a future direction.

\subsection{Pre-training on the partial dataset}

\begin{figure}[t!]
\centering
\includegraphics[width=0.49\textwidth]{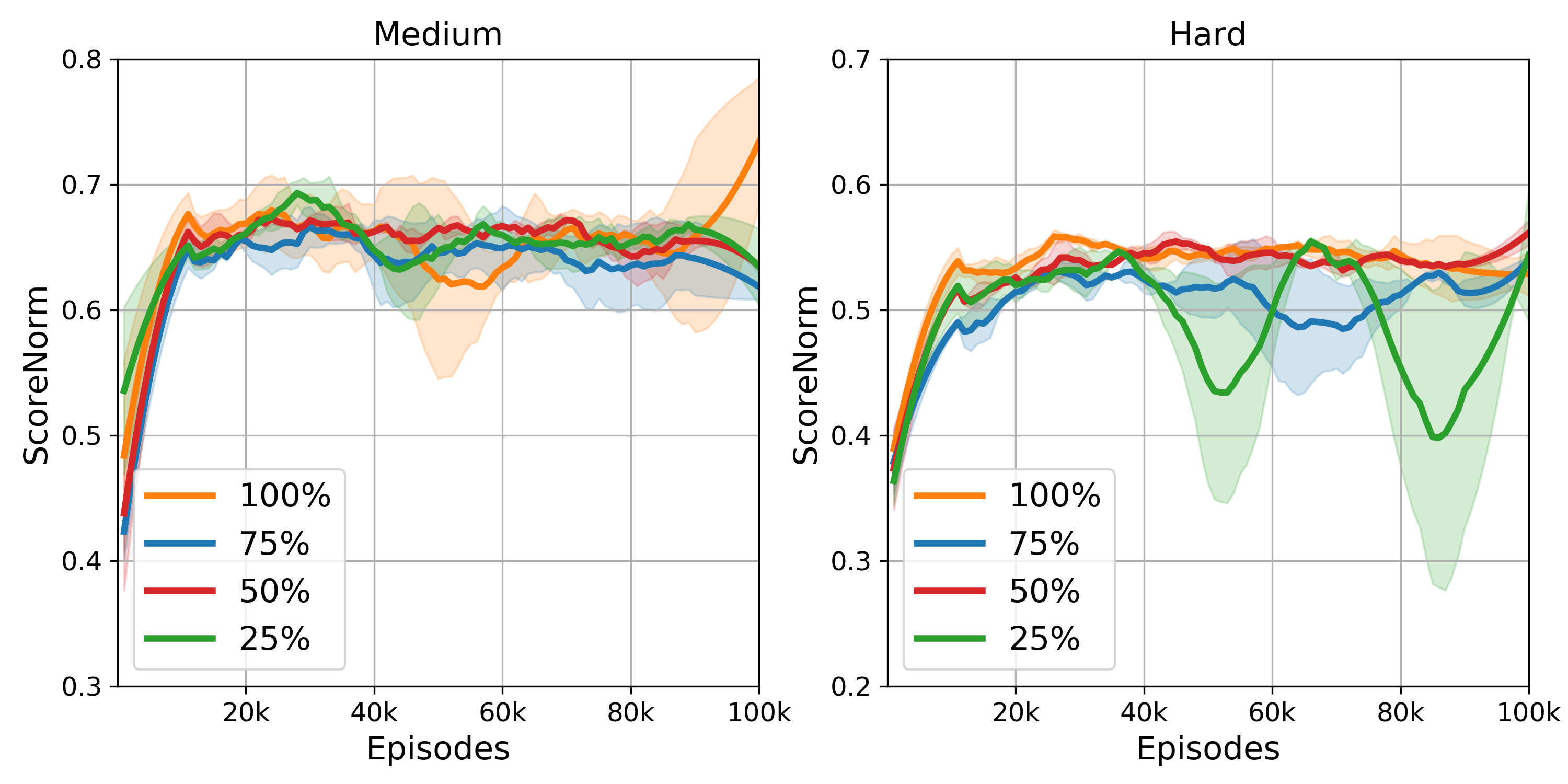}
\caption{The performance of our model with varying amounts of pre-training data.}
\label{exp_4_partial}
\end{figure}

Although we only collect labeled data from the simple games, it is still burdensome for human players to go through the games and answer the questions. 
We are thus interested in investigating how the performance of our QWA (or world-perceiving modules) varies with respect to a reduced amount of pre-training data.
Fig. \ref{exp_4_partial} shows the results, where the pre-training dataset has been reduced to 75\%, 50\% and 25\%, respectively. 
Our model still performs well when the pre-training data is reduced to 75\% and 50\%. When we only use 25\% of the pre-training data, the model exhibits instability during the learning of hard games.
Being pre-trained on a largely-reduced dataset, the world-perceiving modules might be more likely to make wrong predictions with the progress of RL training, leading to the performance fluctuation. However, the final performance of this variant is still comparable. To summarize, our model is robust to limited pre-training data and largely alleviates the burden of human annotations.

\section{Conclusion}

In this paper, we addressed the challenges of low sample efficiency and large action space for deep reinforcement learning in solving text-based games.
We introduced the world-perceiving modules, which are capable of automatic task decomposition and action pruning through answering questions about the environment. 
We proposed a two-phase training framework, which decouples the language learning from the reinforcement learning.
Experimental results show that our method achieves improved performance with high sample efficiency. Besides, it shows robustness against compound error and limited pre-training data. Regarding the future work, we would like to further improve the pre-training performance by introducing contrastive learning objective~\cite{you2020graphcontrastive} and KG-based data augmentation~\cite{zhao2021augmentation}. 

\section*{Acknowledgements}

This work is supported in part by ARC DP21010347, ARC DP180100966 and Facebook Research. Joey Tianyi Zhou is supported by A*STAR SERC Central Research Fund (Use-inspired Basic Research). We thank the anonymous reviewers for their constructive suggestions. We thank Smashicons and Trazobanana for providing the icons in Fig. \ref{arch_1_overview}.

\balance

\bibliographystyle{acl_natbib}

\clearpage
\onecolumn
\appendix

\section*{Appendix}

The appendix is organized as follows: 
Sec. \ref{section_appendix_environment} details the environment. 
Sec. \ref{section_appendix_dataset} illustrates the process for constructing the pre-training datasets. 
Sec. \ref{section_appendix_baseline_details} demonstrates the baselines' architecture and training details.
Sec. \ref{section_appendix_more_result} provides more experimental results.

\section{Game Environment \label{section_appendix_environment}}

In the cooking game~\cite{cote2018textworld}, the player is located in a house, which contains multiple rooms and interactable objects (food, tools, etc.). Her / his task is to follow the recipe to prepare the meal. Each game instance has a unique recipe, including different numbers of ingredients (food objects that are necessary for preparing the meal) and their corresponding preparation requirements (e.g., ``slice'', ``fry''). Besides the textual observation, the KG-based observation can also be directly obtained from the environment. The game sets used in our work contains a task set $\mathcal{T}$ of 268 subtasks, and an action set $\mathcal{A}$ of 1304 actions. Following GATA's experiment setting~\cite{adhikari2020gatav2}, we simplify the game environment by making the action set changeable over time, which can be provided by the TextWorld platform. Note that although the action space is reduced, it still remains challenging as the agent may encounter unseen action candidates~\cite{chandak2019learning,chandak2020lifelong}. We then use a similar way to obtain a changeable task set, which is a combination of the verb set $\{\text{chop, dice, slice, fry, make, get, grill, roast}\}$ and the ingredient set, where the construction details are provided in Appendix \ref{section_appendix_dataset}. Table \ref{table_supl_game_example_simple_medium} and Table \ref{table_supl_game_example_hard} show the KG-based observations $o_t$, corresponding subtask candidates $\mathcal{T}$ and action candidates $\mathcal{A}$. Table \ref{table_supl_subtask_candidates} and Table \ref{table_supl_action_candidates} show more examples of subtasks and actions, respectively. The underlined subtask candidates denote the available subtask set $\mathcal{T}_t$. The underlined action candidates in Table \ref{table_supl_action_candidates} denote the refined action set $\mathcal{A}_t$ after selecting the subtask ``roast carrot''. We still denote the subtask candidate set (action candidate set) as $\mathcal{T}$ ($\mathcal{A}$) to distinguish it from the available subtask set $\mathcal{T}_t$ (refined action set $\mathcal{A}_t$).

\begin{table*}[t!]
\caption{The observations $o_t$, subtask candidates $\mathcal{T}$ and action candidates $\mathcal{A}$ of a simple game and a medium game. The underlined subtask candidates denote the available subtask set $\mathcal{T}_t$.  \label{table_supl_game_example_simple_medium}}
\centering
\resizebox{0.98\textwidth}{!}{
\begin{tabular}{l|p{0.55\textwidth}|p{0.3\textwidth}|p{0.3\textwidth}}
\hline 
\textbf{Game} & \textbf{KG-based observation} & \textbf{Subtask candidates} & \textbf{Action candidates} \\
\hline
\textbf{Simple} 
& ["block of cheese", "cookbook", "part\_of"], ["block of cheese", "fried", "needs"], ["block of cheese", "player", "in"], ["block of cheese", "raw", "is"], ["block of cheese", "sliced", "needs"], ["block of cheese", "uncut", "is"], ["cookbook", "counter", "on"], ["counter", "kitchen", "at"], ["fridge", "kitchen", "at"], ["fridge", "open", "is"], ["knife", "counter", "on"], ["oven", "kitchen", "at"], ["player", "kitchen", "at"], ["stove", "kitchen", "at"], ["table", "kitchen", "at"] & 
\underline{"fry block of cheese"}, \underline{"get knife"}, "chop block of cheese", "dice block of cheese", "get block of cheese", "grill block of cheese", "make meal", "roast block of cheese", "slice block of cheese" & 
"close fridge", "cook block of cheese with oven", "cook block of cheese with stove", "drop block of cheese", "eat block of cheese", "insert block of cheese into fridge", "prepare meal", "put block of cheese on counter", "put block of cheese on stove", "put block of cheese on table", "take cookbook from counter", "take knife from counter" \\
\hline
\textbf{Medium} & 
["bathroom", "corridor", "south\_of"], ["bed", "bedroom", "at"], ["bedroom", "livingroom", "north\_of"], ["block of cheese", "cookbook", "part\_of"], ["block of cheese", "diced", "is"], ["block of cheese", "diced", "needs"], ["block of cheese", "fridge", "in"], ["block of cheese", "fried", "is"], ["block of cheese", "fried", "needs"], ["carrot", "fridge", "in"], ["carrot", "raw", "is"], ["carrot", "uncut", "is"], ["cookbook", "counter", "on"], ["corridor", "bathroom", "north\_of"], ["corridor", "kitchen", "east\_of"], ["corridor", "livingroom", "south\_of"], ["counter", "kitchen", "at"], ["flour", "cookbook", "part\_of"], ["flour", "shelf", "on"], ["fridge", "closed", "is"], ["fridge", "kitchen", "at"], ["frosted-glass door", "closed", "is"], ["frosted-glass door", "kitchen", "west\_of"], ["frosted-glass door", "pantry", "east\_of"], ["kitchen", "corridor", "west\_of"], ["knife", "counter", "on"], ["livingroom", "bedroom", "south\_of"], ["livingroom", "corridor", "north\_of"], ["oven", "kitchen", "at"], ["parsley", "fridge", "in"], ["parsley", "uncut", "is"], ["player", "kitchen", "at"], ["pork chop", "chopped", "is"], ["pork chop", "chopped", "needs"], ["pork chop", "cookbook", "part\_of"], ["pork chop", "fridge", "in"], ["pork chop", "fried", "is"], ["pork chop", "fried", "needs"], ["purple potato", "counter", "on"], ["purple potato", "uncut", "is"], ["red apple", "counter", "on"], ["red apple", "raw", "is"], ["red apple", "uncut", "is"], ["red onion", "fridge", "in"], ["red onion", "raw", "is"], ["red onion", "uncut", "is"], ["red potato", "counter", "on"], ["red potato", "uncut", "is"], ["shelf", "pantry", "at"], ["sofa", "livingroom", "at"], ["stove", "kitchen", "at"], ["table", "kitchen", "at"], ["toilet", "bathroom", "at"], ["white onion", "fridge", "in"], ["white onion", "raw", "is"], ["white onion", "uncut", "is"] & 
\underline{"get block of cheese"}, \underline{"get flour"}, \underline{"get pork chop"}, "chop block of cheese", "chop flour", "chop pork chop", "dice block of cheese", "dice flour", "dice pork chop", "fry block of cheese", "fry flour", "fry pork chop", "get knife", "grill block of cheese", "grill flour", "grill pork chop", "make meal", "roast block of cheese", "roast flour", "roast pork chop", "slice block of cheese", "slice flour", "slice pork chop" & 
"go east", "open fridge", "open frosted-glass door", "take cookbook from counter", "take knife from counter", "take purple potato from counter", "take red apple from counter", "take red potato from counter" \\
\hline
\end{tabular}
}
\end{table*}

\begin{table*}[t!]
\caption{The observations $o_t$, subtask candidates $\mathcal{T}$ and action candidates $\mathcal{A}$ of a hard game. The underlined subtask candidates denote the available subtask set $\mathcal{T}_t$. The underlined action candidates denote the refined action set $\mathcal{A}_t$ after selecting the subtask ``roast carrot''.
\label{table_supl_game_example_hard}}
\centering
\resizebox{0.98\textwidth}{!}{
\begin{tabular}{l|p{0.55\textwidth}|p{0.3\textwidth}|p{0.3\textwidth}}
\hline 
\textbf{Game} & \textbf{KG-based observation} & \textbf{Subtask candidates} & \textbf{Action candidates} \\
\hline
\textbf{Hard} & 
["backyard", "garden", "west\_of"], ["barn door", "backyard", "west\_of"], ["barn door", "closed", "is"], ["barn door", "shed", "east\_of"], ["bathroom", "corridor", "east\_of"], ["bbq", "backyard", "at"], ["bed", "bedroom", "at"], ["bedroom", "corridor", "north\_of"], ["bedroom", "livingroom", "south\_of"], ["carrot", "cookbook", "part\_of"], ["carrot", "player", "in"], ["carrot", "raw", "is"], ["carrot", "roasted", "needs"], ["carrot", "sliced", "needs"],["carrot", "uncut", "is"], ["commercial glass door", "closed", "is"], ["commercial glass door", "street", "east\_of"], ["commercial glass door", "supermarket", "west\_of"], ["cookbook", "table", "on"], ["corridor", "bathroom", "west\_of"], ["corridor", "bedroom", "south\_of"], ["counter", "kitchen", "at"], ["driveway", "street", "north\_of"], ["fridge", "closed", "is"], ["fridge", "kitchen", "at"], ["front door", "closed", "is"], ["front door", "driveway", "west\_of"], ["front door", "livingroom", "east\_of"], ["frosted-glass door", "closed", "is"], ["frosted-glass door", "kitchen", "south\_of"], ["frosted-glass door", "pantry", "north\_of"], ["garden", "backyard", "east\_of"], ["kitchen", "livingroom", "west\_of"], ["knife", "counter", "on"], ["livingroom", "bedroom", "north\_of"], ["livingroom", "kitchen", "east\_of"], ["oven", "kitchen", "at"], ["patio chair", "backyard", "at"], ["patio door", "backyard", "north\_of"], ["patio door", "corridor", "south\_of"], ["patio door", "open", "is"], ["patio table", "backyard", "at"], ["player", "backyard", "at"], ["red apple", "counter", "on"], ["red apple", "raw", "is"], ["red apple", "uncut", "is"], ["red hot pepper", "cookbook", "part\_of"], ["red hot pepper", "player", "in"], ["red hot pepper", "raw", "is"], ["red hot pepper", "roasted", "needs"], ["red hot pepper", "sliced", "needs"], ["red hot pepper", "uncut", "is"], ["red onion", "garden", "at"], ["red onion", "raw", "is"], ["red onion", "uncut", "is"], ["shelf", "pantry", "at"], ["showcase", "supermarket", "at"], ["sofa", "livingroom", "at"], ["stove", "kitchen", "at"], ["street", "driveway", "south\_of"], ["table", "kitchen", "at"], ["toilet", "bathroom", "at"], ["toolbox", "closed", "is"], ["toolbox", "shed", "at"], ["white onion", "chopped", "needs"], ["white onion", "cookbook", "part\_of"], ["white onion", "grilled", "needs"], ["white onion", "player", "in"], ["white onion", "raw", "is"], ["white onion", "uncut", "is"], ["workbench", "shed", "at"], ["yellow bell pepper", "garden", "at"], ["yellow bell pepper", "raw", "is"], ["yellow bell pepper", "uncut", "is"] &

\underline{"roast carrot"}, \underline{"roast red hot pepper"}, \underline{"grill white onion"}, \underline{"get knife"}, "chop carrot", "chop red hot pepper", "chop white onion", "dice carrot", "dice red hot pepper", "dice white onion", "fry carrot", "fry red hot pepper", "fry white onion", "get carrot", "get red hot pepper", "get white onion", "grill carrot", "grill red hot pepper", "make meal",  "roast white onion", "slice carrot", "slice red hot pepper", "slice white onion" & 

\underline{"go east"}, \underline{"go north"}, \underline{"open barn door"}, \underline{"open patio door"}, "close patio door", "cook carrot with bbq", "cook red hot pepper with bbq", "cook white onion with bbq", "drop carrot", "drop red hot pepper", "drop white onion", "eat carrot", "eat red hot pepper", "eat white onion", "put carrot on patio chair", "put carrot on patio table", "put red hot pepper on patio chair", "put red hot pepper on patio table", "put white onion on patio chair", "put white onion on patio table" \\
\hline
\end{tabular}
}
\end{table*}

\begin{table*}[t!]
\caption{Examples of subtasks. \label{table_supl_subtask_candidates}}
\centering
\resizebox{0.98\textwidth}{!}{
\begin{tabular}{p{0.35\textwidth}|p{0.35\textwidth}|p{0.35\textwidth}}
\hline 
\multicolumn{3}{c}{\textbf{Subtask candidates}} \\
\hline
chop banana & chop black pepper & chop block of cheese \\
chop olive oil & chop orange bell pepper & chop parsley \\
chop vegetable oil & chop water & chop white onion \\
dice cilantro & dice egg & dice flour \\
dice red bell pepper & dice red hot pepper & dice red onion \\
dice yellow potato & fry banana & fry black pepper \\
fry milk & fry olive oil & fry orange bell pepper \\
fry tomato & fry vegetable oil & fry water \\
get chicken wing & get cilantro & get egg \\
get purple potato & get red apple & get red bell pepper \\
get yellow bell pepper & get yellow onion & get yellow potato \\
grill green hot pepper & grill lettuce & grill milk \\
grill salt & grill sugar & grill tomato \\
roast carrot & roast chicken breast & roast chicken leg \\
roast peanut oil & roast pork chop & roast purple potato \\
roast white tuna & roast yellow apple & roast yellow bell pepper \\
slice green apple & slice green bell pepper & slice green hot pepper \\
slice red potato & slice red tuna & slice salt \\
\hline
\end{tabular}
}
\end{table*}

\begin{table*}[t!]
\caption{Examples of actions. \label{table_supl_action_candidates}}
\centering
\resizebox{0.98\textwidth}{!}{
\begin{tabular}{p{0.35\textwidth}|p{0.35\textwidth}|p{0.35\textwidth}}
\hline 
\multicolumn{3}{c}{\textbf{Action candidates}} \\
\hline
chop banana with knife & chop block of cheese with knife & chop carrot with knife \\
cook block of cheese with oven & cook block of cheese with stove & cook carrot with bbq \\
cook orange bell pepper with oven & cook orange bell pepper with stove & cook parsley with bbq \\
cook water with stove & cook white onion with bbq & cook white onion with oven \\
drink water & drop banana & drop black pepper \\
eat carrot & eat chicken breast & eat chicken leg \\
insert block of cheese into toolbox & insert carrot into fridge & insert carrot into toolbox \\
insert red onion into fridge & insert red onion into toolbox & insert red potato into fridge \\
put banana on shelf & put banana on showcase & put banana on sofa \\
put chicken breast on showcase & put chicken breast on sofa & put chicken breast on stove \\
put egg on patio table & put egg on shelf & put egg on showcase \\
put green hot pepper on shelf & put green hot pepper on showcase & put green hot pepper on sofa \\
put olive oil on patio chair & put olive oil on patio table & put olive oil on shelf \\
put pork chop on sofa & put pork chop on stove & put pork chop on table \\
put red hot pepper on table & put red hot pepper on toilet & put red hot pepper on workbench \\
put salt on workbench & put sugar on bed & put sugar on counter \\
put white onion on shelf & put white onion on showcase & put white onion on sofa \\
put yellow onion on sofa & put yellow onion on stove & put yellow onion on table \\
take banana from patio chair & take banana from patio table & take banana from shelf \\
take carrot from showcase & take carrot from sofa & take carrot from stove \\
take chicken wing from toolbox & take chicken wing from workbench & take cilantro \\
take green apple from bed & take green apple from counter & take green apple from fridge \\
take lettuce from sofa & take lettuce from stove & take lettuce from table \\
take orange bell pepper from workbench & take parsley & take parsley from bed \\
take purple potato from showcase & take purple potato from sofa & take purple potato from stove \\
take red hot pepper from toolbox & take red hot pepper from workbench & take red onion \\
take salt from counter & take salt from fridge & take salt from patio chair \\
take water from counter & take water from fridge & take water from patio chair \\
take yellow apple from sofa & take yellow apple from stove & take yellow apple from table \\
\hline
\end{tabular}
}
\end{table*}

\clearpage

\section{Pre-training Datasets \label{section_appendix_dataset}}

\begin{figure}[h]
\centering
\includegraphics[width=1.0\textwidth]{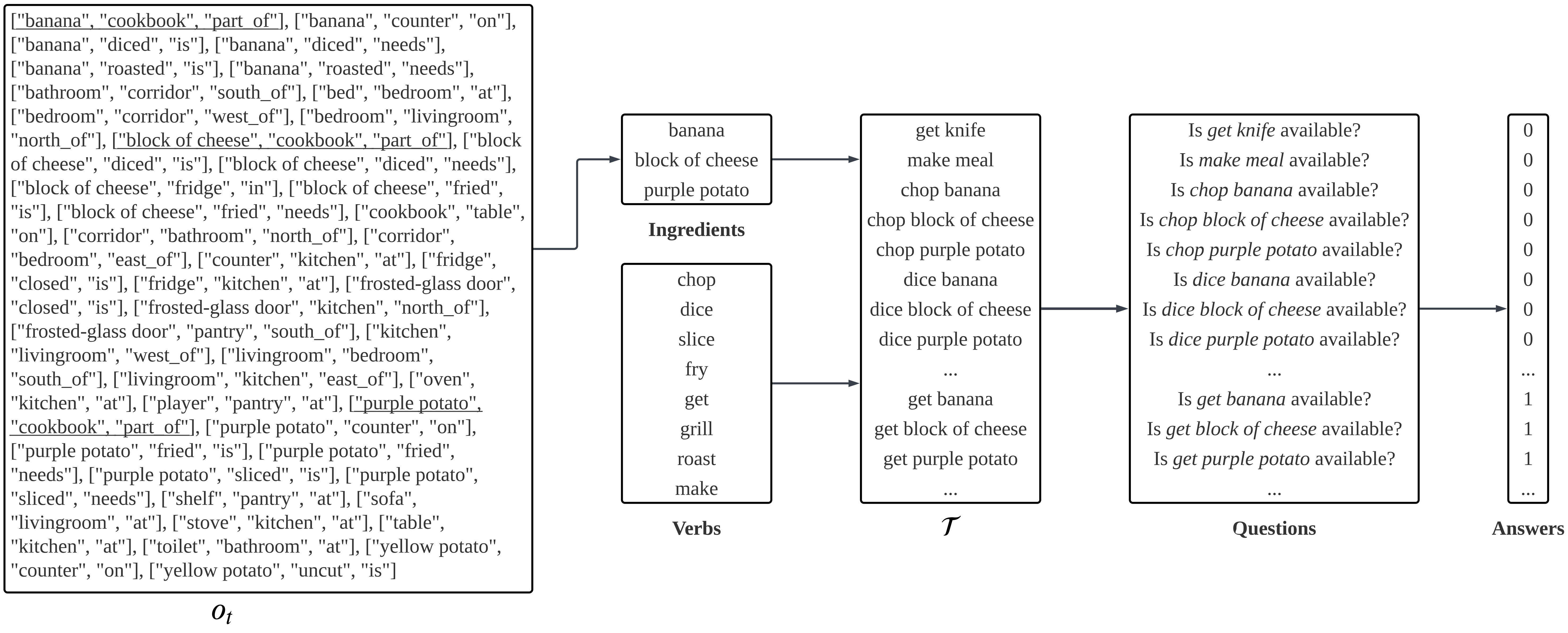}
\caption{The construction process of the subtask set $\mathcal{T}$, and the pre-training dataset for task decomposition.}
\label{arch_5_sup_VTdataset}
\end{figure}

\begin{figure}[h]
\centering
\includegraphics[width=0.8\textwidth]{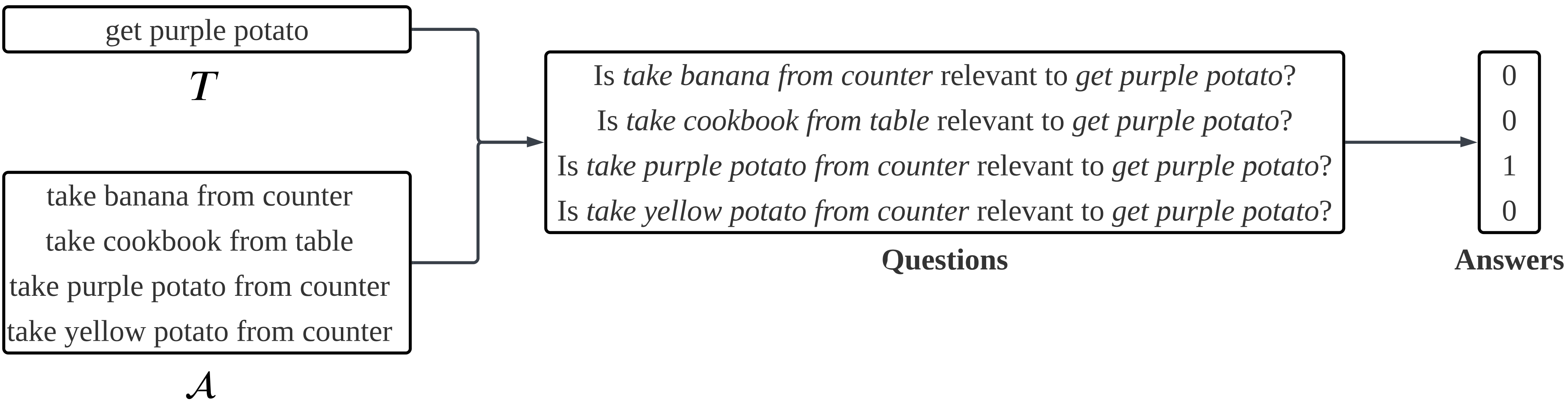}
\caption{The construction process of the pre-training dataset for action pruning.}
\label{arch_5_sup_VAdataset}
\end{figure}

We build separate datasets for each pre-training task (task decomposition, action pruning, and imitation learning). We first let the player to go through each simple game, then construct the datasets upon the interaction data. 
For each time step, the game environment provides the player with the action set $\mathcal{A}$ and the KG-based observation $o_t$, which is represented as a set of triplets. 
We use a simple method to build the subtask set $\mathcal{T}$ from $o_t$: 
As shown in Fig. \ref{arch_5_sup_VTdataset}, we first obtain the ingredients by extracting the nodes having the relation ``part\_of'' with the node ``cookbook''. Then we build $\mathcal{T}$ as the Cartesian product of the ingredients and the verbs $\{\text{chop, dice, slice, fry, get, grill, roast}\}$ plus two special subtasks ``get knife'' and ``make meal''. 
The player is required to select a subtask $T_t \in \mathcal{T}$, and select an action $a_t \in \mathcal{A}$. 
After executing $a_t$, the environment will transit to next state $s_{t+1}$, and the player will receive $o_{t+1}$ and $r_{t+1}$ to form a transition $\{o_t, \mathcal{T}, T_t, \mathcal{A}, a_t, o_{t+1}, r_{t+1}\}$, where $\{o_t, \mathcal{T}, T_t, \mathcal{A}, a_t\}$ will be used for imitation learning. 
Fig. \ref{arch_5_sup_VTdataset} shows the construction process of the pre-training dataset for task decomposition. Each subtask candidate $T \in \mathcal{T}$ will formulate a question ``Is $T$ available?'', whose answer is 1 (yes) if $T$ is an available subtask for $o_t$, otherwise 0 (no). 
Fig. \ref{arch_5_sup_VAdataset} shows the construction process of the pre-training dataset for action pruning. 
The action selector is made invariant of $o_t$, that we consider every subtask candidate $T \in \mathcal{T}$ during pre-training, regardless of whether $T$ is a currently-available subtask. Each action candidate $a \in \mathcal{A}$ will be paired with $T$ to formulate a question ``Is $a$ relevant to $T$'', whose answer is 1 if $a$ is relevant to $T$, otherwise 0. 

\clearpage

\section{Baseline details \label{section_appendix_baseline_details}}

\subsection{GATA}

\begin{figure}[h]
\centering
\includegraphics[width=0.6\textwidth]{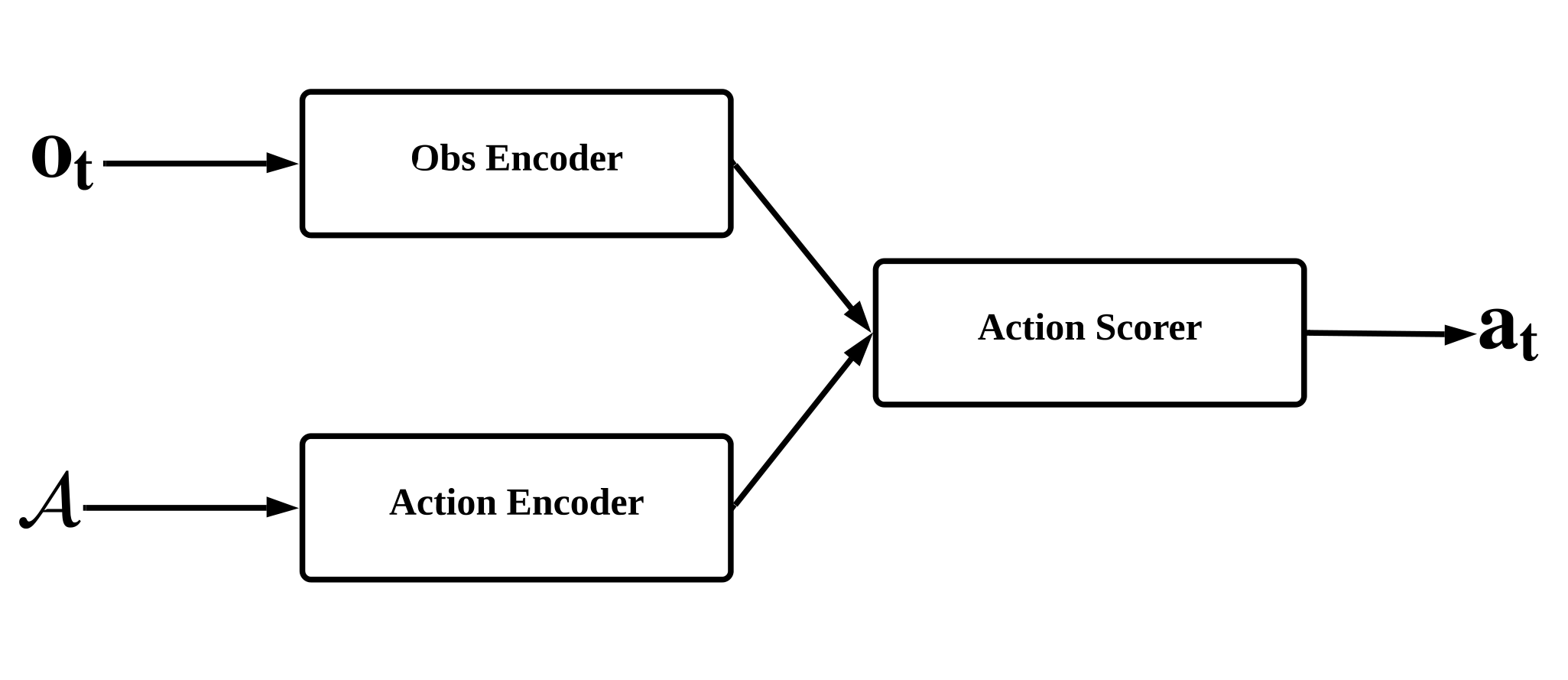}
\caption{The architecture of GATA baseline.}
\label{arch_4_sup_GATA}
\end{figure}

Fig. \ref{arch_4_sup_GATA} shows our backbone model GATA, which consists of an observation encoder, an action encoder and an action scorer. The observation encoder is a graph encoder for encoding the KG-based observation $o_t$, and the action encoder is a text encoder to encode the action set $\mathcal{A}$ as a stack of action candidate representations. The observation representation will be paired with each action candidate, and then fed into the action scorer, which consists of linear layers. 

\begin{figure}[h]
\centering
\includegraphics[width=0.6\textwidth]{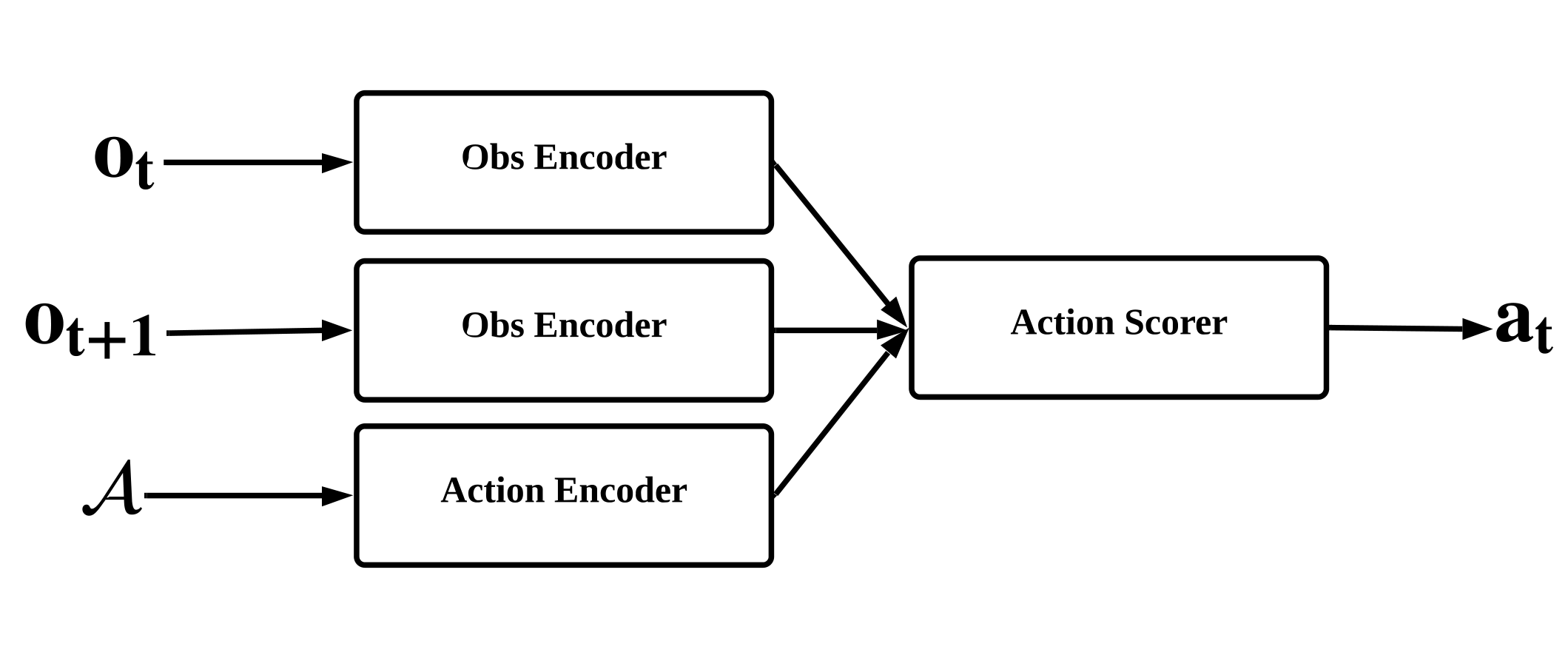}
\caption{The architecture of GATA for action prediction.}
\label{arch_4_sup_GATA_AP}
\end{figure}

We train the GATA through reinforcement learning, the experiment setting is same with Sec. \ref{section_implementation_details}.
Instead of initializing the word embedding, node embedding and edge embedding with fastText word vectors~\cite{mikolov2017fasttext}, we found that the action prediction task (AP), which is also included in GATA's work~\cite{adhikari2020gatav2}, could provide better initialization. 
In light of this, we could like to conduct such task, and apply the AP initialization to all encoders (observation encoder, task encoder, action encoder). 
Fig. \ref{arch_4_sup_GATA_AP} shows the action predicting process. Given the transition data, the task is to predict the action $a_t \in \mathcal{A}$ given the current observation $o_t$, and the next observation $o_{t+1}$ after executing $a_t$. 
The transition data for AP task is collected from the FTWP game set and is provided by GATA's released code.

\subsection{IL}

\begin{figure}[h]
\centering
\includegraphics[width=0.6\textwidth]{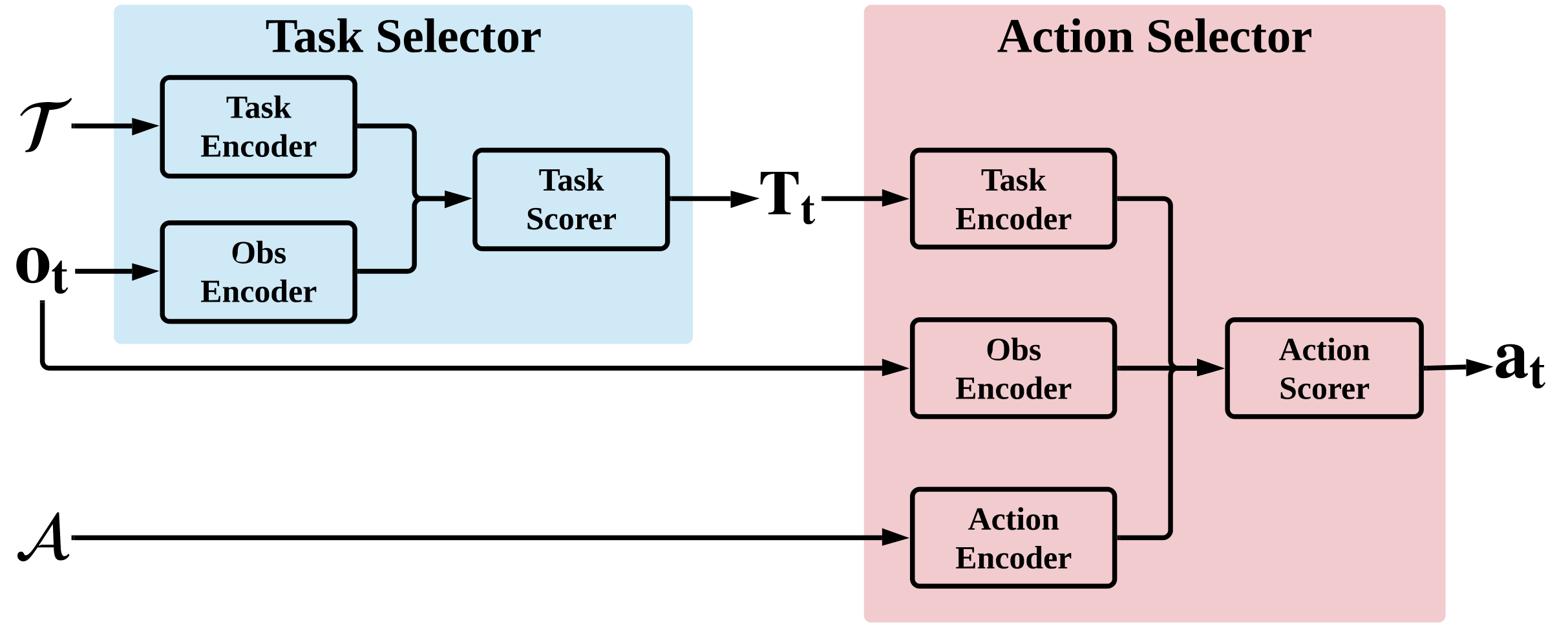}
\caption{The architecture of IL baseline.}
\label{arch_4_sup_IL}
\end{figure}

Fig. \ref{arch_4_sup_IL} shows the IL baseline. We follow ~\cite{chen2021askyourhumans} to conduct a two-phase training process: imitation pre-training and reinforcement fine-tuning. In the imitation pre-training phase, we use the transition data to train both the task selector ($f(o_t, \mathcal{T}) \rightarrow T_t$) and the action selector ($f(o_t, T_t, \mathcal{A}) \rightarrow a_t$) through supervised learning. 
The modules are optimized via cross entropy loss and Adam optimizer with learning rate 0.001. 
We train the modules with batch size 128 for up to 50 epochs. 
Then in the reinforcement fine-tuning phase, we freeze the task selector and fine-tune the action selector through reinforcement learning, where the experiment setting is same with QWA and GATA. 

\clearpage

\section{More experimental results \label{section_appendix_more_result}}

\begin{figure*}[h]
\centering
\includegraphics[width=0.98\textwidth]{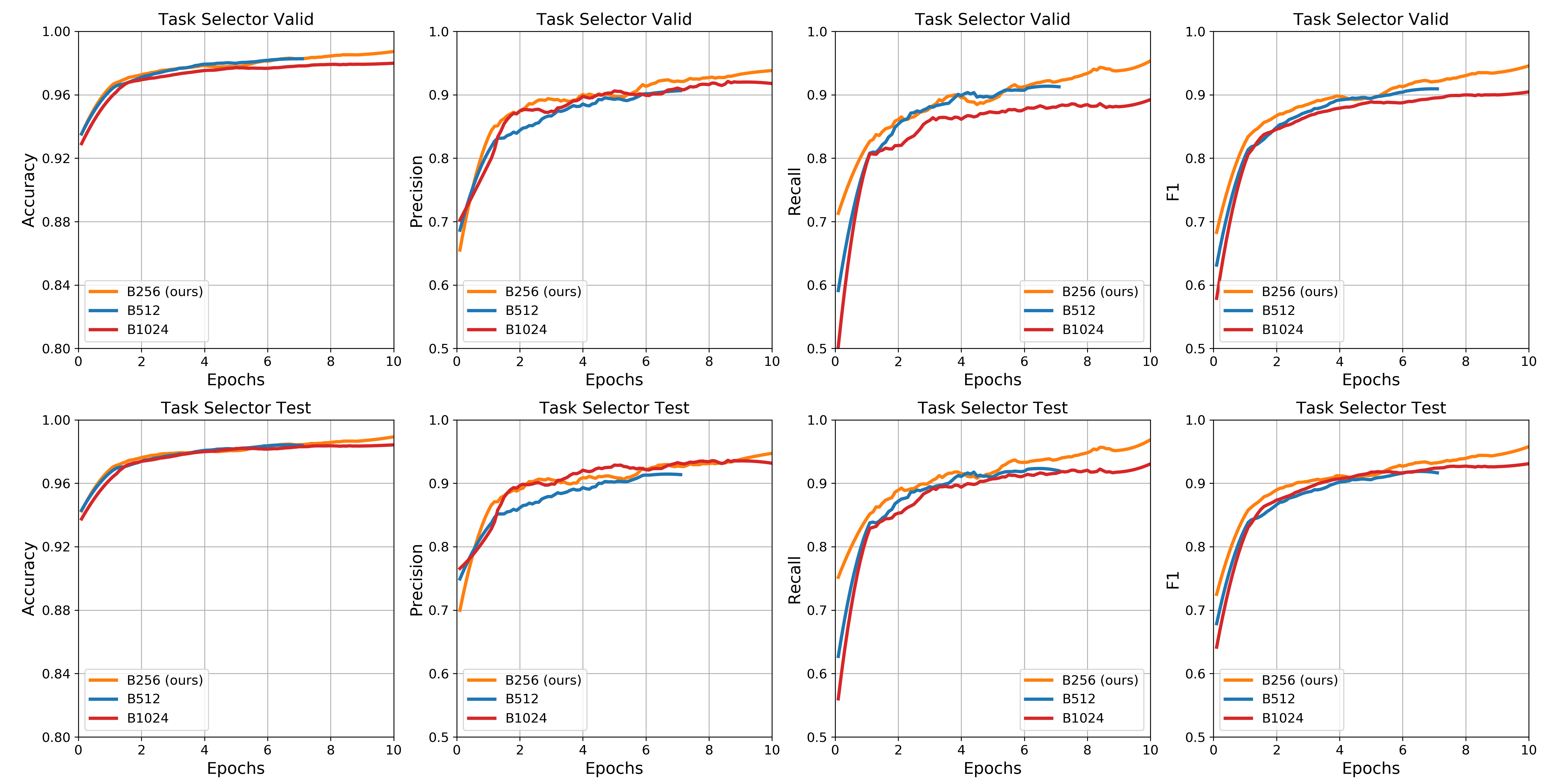}
\caption{The pre-training performance of QWA's task selector. The results are averaged by 3 random seeds, we omit the standard deviation as the performance is relatively stable.}
\label{supl_VTpretrain}
\end{figure*}

\begin{figure*}[h]
\centering
\includegraphics[width=0.98\textwidth]{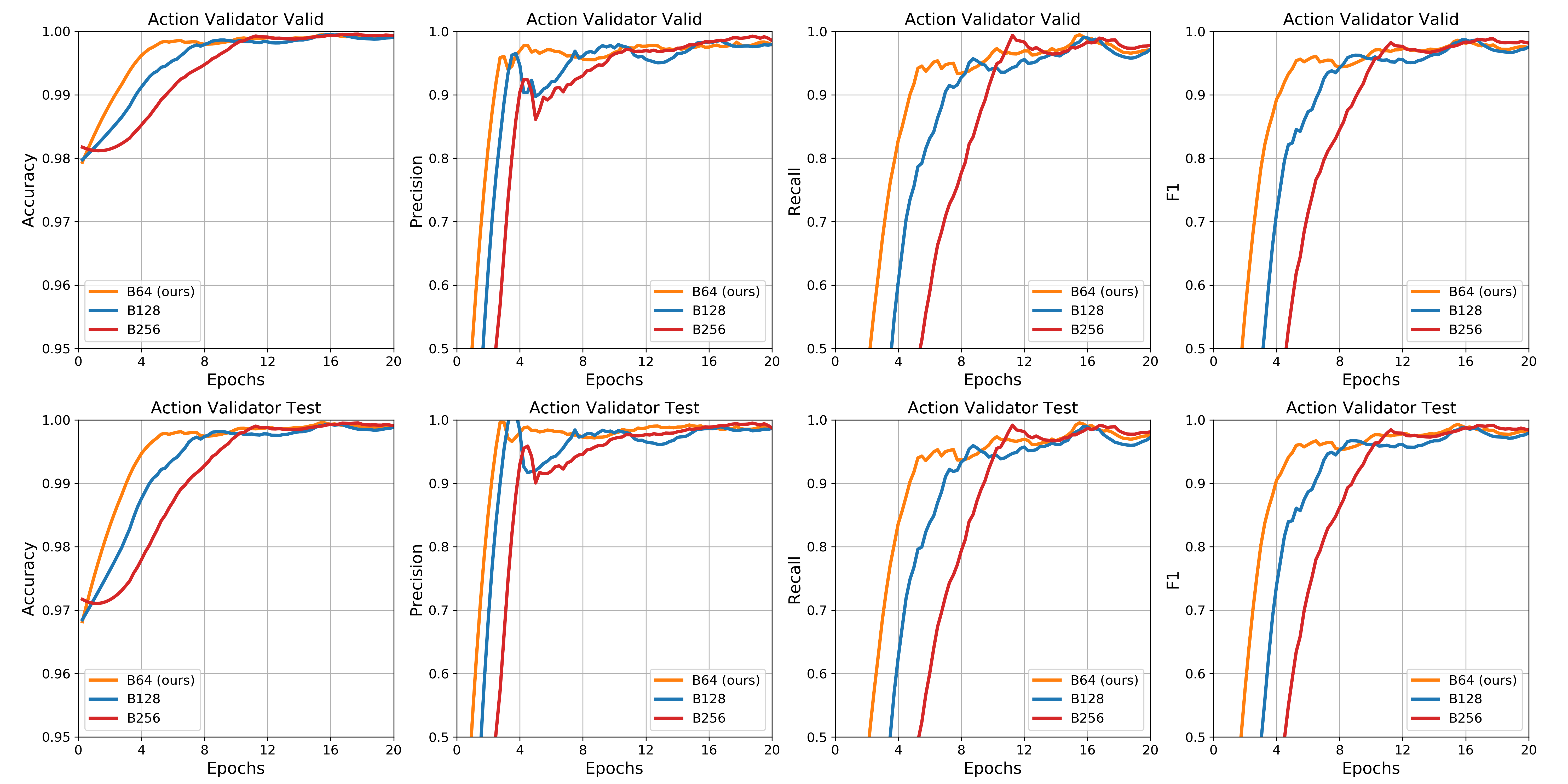}
\caption{The pre-training performance of QWA's action validator. }
\label{supl_VApretrain}
\end{figure*}

\begin{figure*}[t!]
\centering
\includegraphics[width=0.98\textwidth]{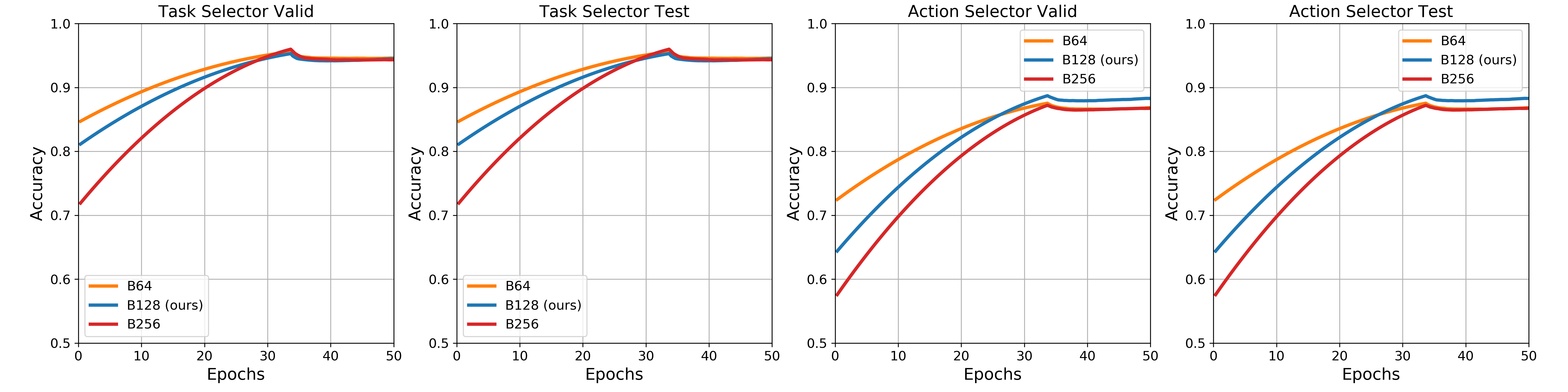}
\caption{The pre-training performance of IL's task selector and action selector.}
\label{supl_ILpretrain}
\end{figure*}

In the pre-training phase, we conduct rough hyper-parameter tuning by varying batch sizes. 
Fig. \ref{supl_VTpretrain} and Fig. \ref{supl_VApretrain} show the pre-training performance of QWA's task selector and action validator, respectively. Fig. \ref{supl_ILpretrain} shows the pre-training performance of IL baseline. 

\begin{figure*}[t!]
\centering
\includegraphics[width=0.98\textwidth]{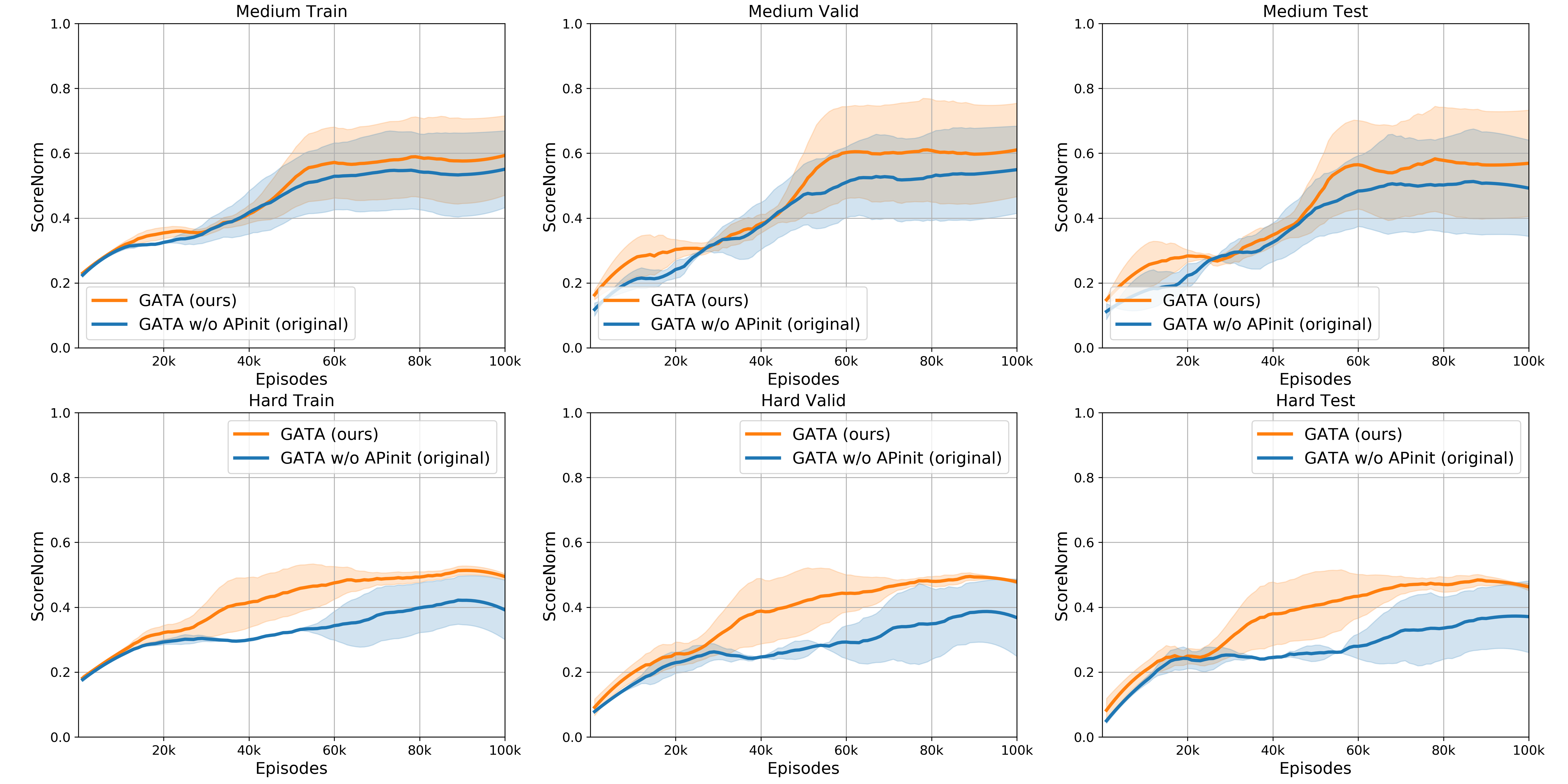}
\caption{The RL performance of our GATA baseline and the original GATA without AP initialization.}
\label{exp_supl_0_APinit}
\end{figure*}

\begin{figure*}[t!]
\centering
\includegraphics[width=0.98\textwidth]{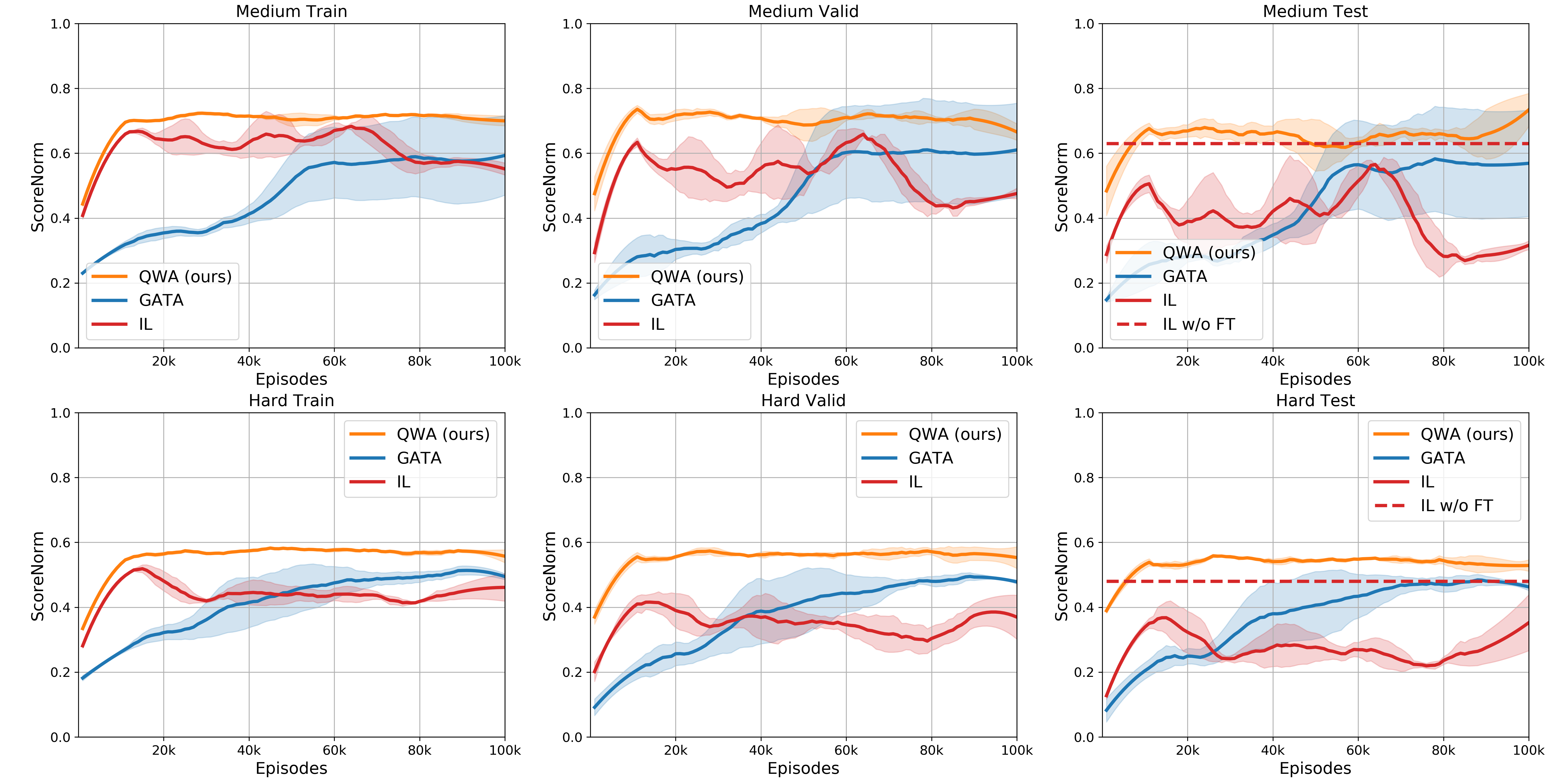}
\caption{The RL performance of models with respect to training episodes (the full result of Fig. \ref{exp_1_main}).}
\label{exp_supl_1_main}
\end{figure*}

\begin{figure*}[t!]
\centering
\includegraphics[width=0.98\textwidth]{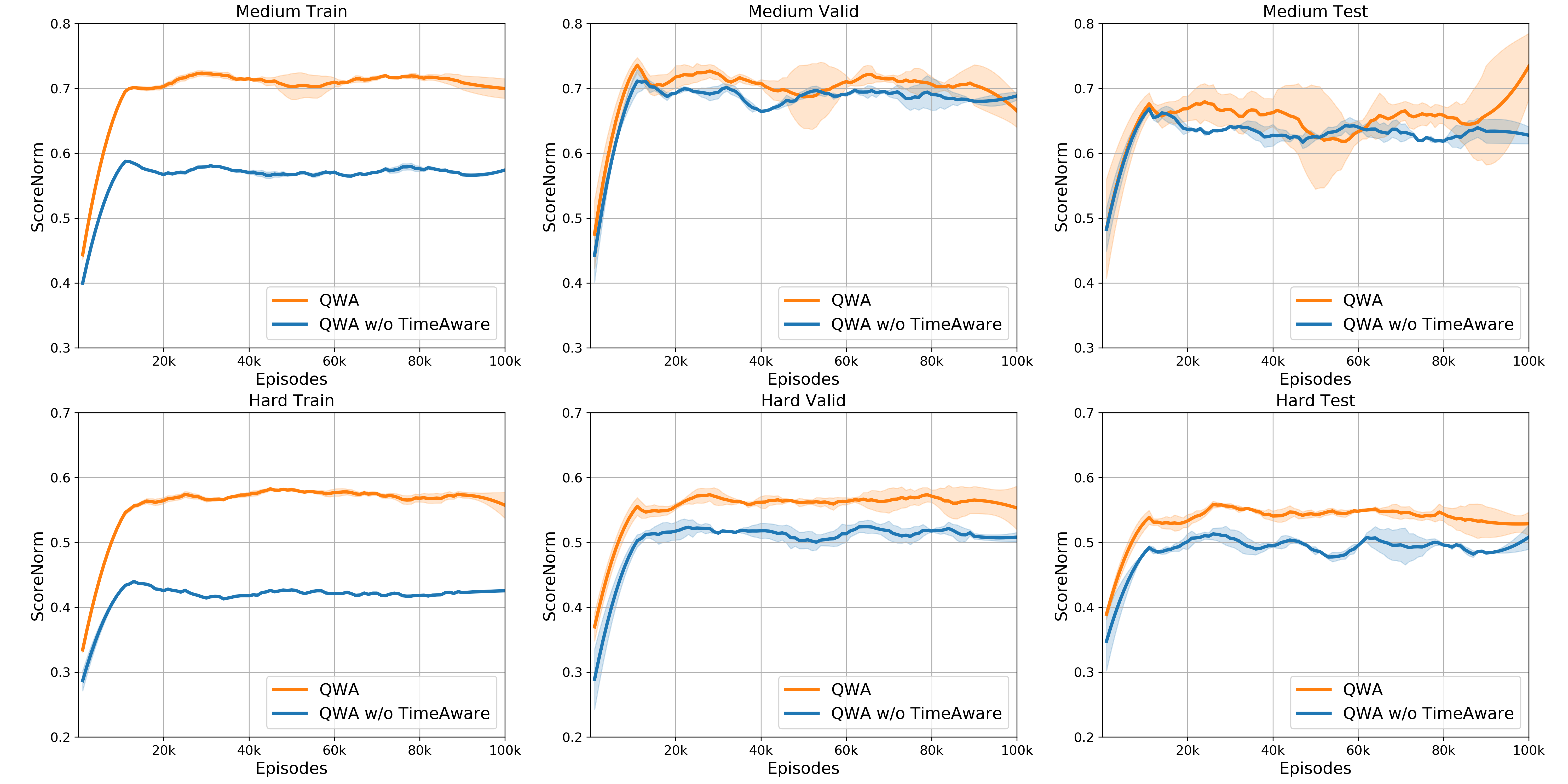}
\caption{The RL performance of our model and the variant without time-awareness (the full result of Fig. \ref{exp_2_ablationTimeAware}). }
\label{exp_supl_2_ablationTimeAware}
\end{figure*}

\begin{figure*}[t!]
\centering
\includegraphics[width=0.98\textwidth]{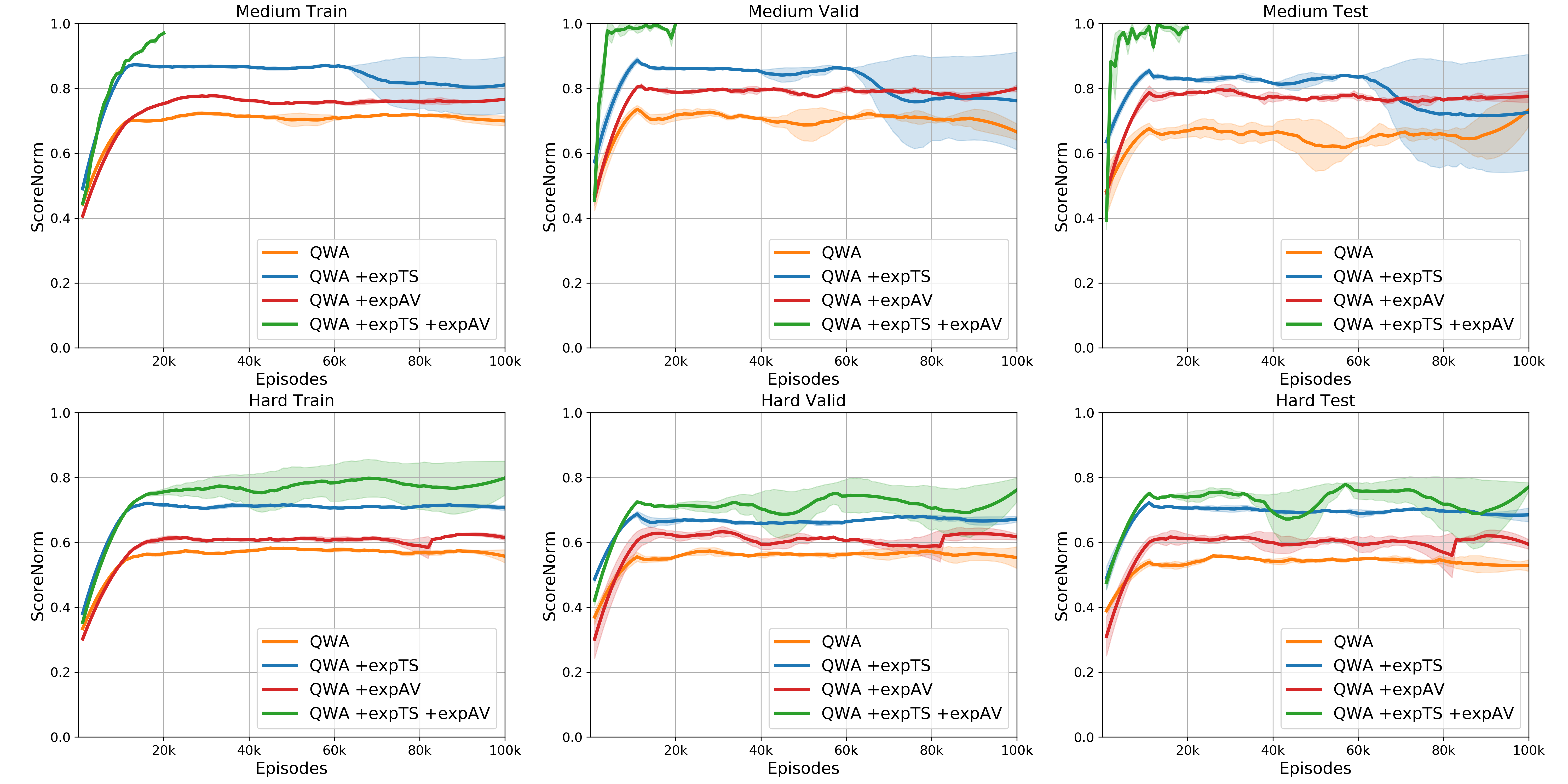}
\caption{The performance of our model and the variants with expert modules (the full result of Fig. \ref{exp_2_ablationExpert}). }
\label{exp_supl_2_ablationExpert}
\end{figure*}

\begin{figure*}[t!]
\centering
\includegraphics[width=0.98\textwidth]{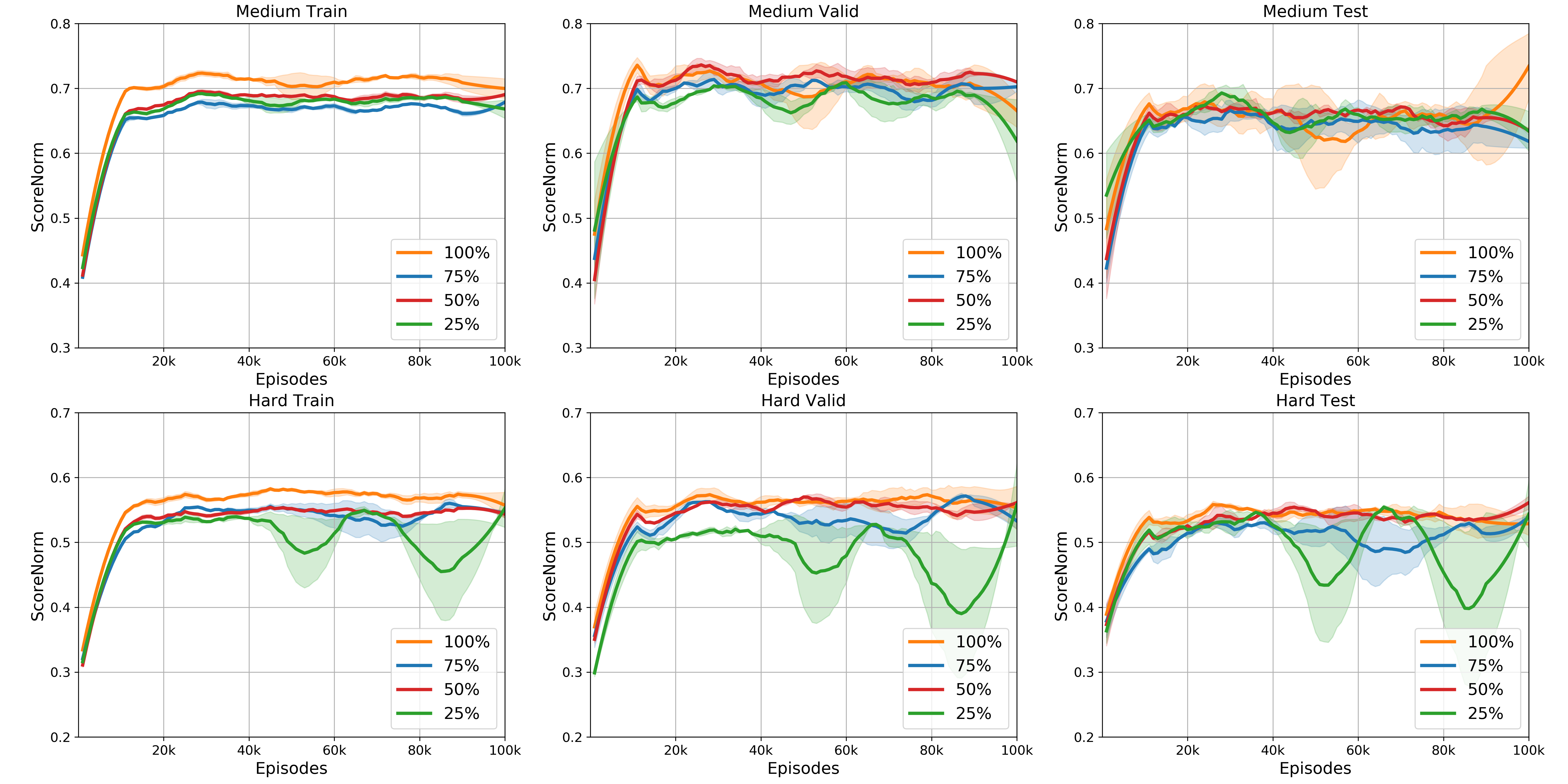}
\caption{The performance of our model with varying amounts of pre-training data (the full result of Fig. \ref{exp_4_partial}). }
\label{exp_supl_4_partial}
\end{figure*}

Fig. \ref{exp_supl_0_APinit} compares our GATA and the original GATA without the action prediction initialization. Fig. \ref{exp_supl_1_main}, Fig. \ref{exp_supl_2_ablationTimeAware}, Fig. \ref{exp_supl_2_ablationExpert} and Fig. \ref{exp_supl_4_partial} show the full results of Fig. \ref{exp_1_main}, Fig. \ref{exp_2_ablationTimeAware}, Fig. \ref{exp_2_ablationExpert} and Fig. \ref{exp_4_partial}, respectively.

\end{document}